\documentclass[preprint,12pt,3p]{elsarticle}
\usepackage{float}
\floatstyle{plaintop}
\restylefloat{table}
\restylefloat{figure}

\usepackage{hyperref}
\hypersetup{colorlinks,allcolors=black}
\usepackage{amsmath,amssymb,amsfonts}
\usepackage{graphicx}
\usepackage[table,xcdraw]{xcolor}
\usepackage{graphicx}

\usepackage{csvsimple}

\journal{IEEE}

\begin{document}

\begin{frontmatter}
\title{Diagnosis of Schizophrenia: A comprehensive evaluation}

\author[label1]{M. Tanveer} 
\address[label1]{Department of Mathematics, Indian Institute of Technology Indore, Simrol, Indore, 453552, India}
\ead{mtanveer@iiti.ac.in}

\author[label2]{Jatin Jangir}
\address[label2]{Department of Electrical Engineering, Indian Institute of Technology Indore, Simrol, Indore, 453552, India}
\ead{ee180002022@iiti.ac.in}

\author[label1]{M.A. Ganaie}
\ead{phd1901141006@iiti.ac.in}

\author[label3]{Iman Beheshti}
\address[label3]{Department of Human Anatomy and Cell Science, Rady Faculty of Health Sciences, Max Rady College of Medicine, University of Manitoba, Winnipeg, MB, Canada.}
\ead{Iman.beheshti@umanitoba.ca}

\author[label1]{M. Tabish}
\ead{msc1903141002@iiti.ac.in}

\author[label2]{Nikunj Chhabra}
\ead{ee180002040@iiti.ac.in}

\begin{abstract}
Machine learning models have been successfully employed in the diagnosis of Schizophrenia disease. The impact of classification models and the feature selection techniques on the diagnosis of Schizophrenia have not been evaluated. Here, we sought to access the performance of classification models along with different feature selection approaches on the structural magnetic resonance imaging data. The data consist of $72$ subjects with Schizophrenia and $74$ healthy control subjects. We evaluated different classification algorithms based on support vector machine (SVM), random forest, kernel ridge regression and randomized neural networks. Moreover, we evaluated T-Test, Receiver Operator Characteristics (ROC), Wilcoxon, entropy, Bhattacharyya, Minimum Redundancy Maximum Relevance (MRMR) and Neighbourhood Component Analysis (NCA) as the feature selection techniques. Based on the evaluation, SVM based models with Gaussian kernel proved better compared to other classification models and Wilcoxon feature selection emerged as the best feature selection approach. Moreover, in terms of data modality the performance on integration of the grey matter and white matter proved better compared to the performance  on the grey and white matter individually. Our evaluation showed that classification algorithms along with the feature selection approaches impact the diagnosis of Schizophrenia disease. This indicates that proper selection of the features and the classification models can improve the diagnosis of Schizophrenia.
\end{abstract}

\begin{keyword}
 Schizophrenia, classification, machine learning.
\end{keyword}
\end{frontmatter}

%


\vspace{-4mm}
\section{Introduction}
Schizophrenia is a severe mental disorder that affects millions of people worldwide. Schizophrenia makes people slowly lose contact with reality, leading to hallucinations, delusions, and extremely disordered thinking. Patients report hearing voices or seeing things that are not there; they also tend to develop fixed and false beliefs. Suicidal tendency is also a common trait among Schizophrenia patients. Moreover, patients inflicted with Schizophrenia are $2-3$ times more likely to die than the general public due to patients not seeking aid for preventable physical diseases \cite{pmid24313570}. Luckily, Schizophrenia is treatable with medicines and psycho-social support, and these methods have proven successful \cite{WHOSchizophrenia}. Thus, the central blockade in eradicating Schizophrenia is lack of its early detection.

Several attempts have been made to remedy this problem. Many studies show promising results but, Machine Learning (ML) has seen little use in clinical practice for Schizophrenia. This can perhaps be credited to the unreliability or stability of some machine learning models; this influences the consensus that ML solutions are not dependable and cannot be trusted, especially for a medical job. Even though doctors sometimes make mistakes themselves, people still trust them. Developing this same level of trust for a machine would be an arduous task. Machine learning should be seen as a research tool to advance the field of study, not the be-all and end-all. In the past few years, extensive research has been done on various classification algorithms and their improved versions have been proposed like for SVM, it's extensions like twin support vector machine (TWSVM) \cite{khemchandani2007twin}, twin bounded SVM (TBSVM) \cite{shao2011improvements} etc. are proposed to improve the performance of SVM algorithm. Other methods such as  $k$-nearest neighbour (KNN) \cite{hand2007principles}, random forest (RaF) \cite{breiman2001random} have also been thoroughly studied. Interested readers can refer to the comprehensive review on TWSVM \cite{tanveer2021comprehensive}.

Our intent with this study is to perform a comprehensive evaluation and bring awareness towards using different classification algorithms and their variants and extensions for diagnosing schizophrenia disease. In this study, we evaluate single-modal methods using exclusively Structural MRI \color{black}(sMRI) scans to train and validate them. We use the same dataset for all the algorithms. The results of this study will help choose a suitable classification algorithm and feature selection technique based on the requirement. The rest of the paper is organized as follows: In Section \ref{Sec:Subjects and Methods}, we discuss about subjects, $3D$ MRI processing and give a brief description of various classification algorithms and feature selection techniques used and also about validation and experimental setup. Performance of various classification algorithms on white matter, grey matter and integrated matter \color{black}is discussed in Section \ref{Sec:Results}. Analysis and summarizing of results is done in Section \ref{Sec:Discussion}. Conclusions and future works are discussed in Section \ref{Sec:Conclusions}. Please note, Figures and Tables referenced from the attached Supplementary Paper have been suffixed with an ``S-".

\section{Related Works}
\label{Sec:related_works}
The current diagnosis scheme for Schizophrenia is to rule out other mental disorders and then employ psychiatric and physical screening. The studies \cite{pmid11343862, steardo2020application} evaluated magnetic resonance imaging (MRI) scans for the detection of Schizophrenia. A
review of MRI findings in schizophrenia \cite{pmid11343862} discusses brain abnormalities due to Schizophrenia. Thus, there is significant evidence that by using MRI scans, one can exploit ML techniques to automate and improve the detection of Schizophrenia. Several studies \cite{guo2020support, zarogianni2013towards} have already attempted to do so with varying degree of success. A basic summary for most of the studies is: process the MRI scan into a usable format, apply a feature extraction algorithm on the MRI scan to select the appropriate features, then finally use a classification algorithm for the diagnosis of schizophrenia.

The classification of schizophrenia patients and healthy controls from sMRI scans in two large independent samples was studied in \cite{Nieuwenhuis2012}. The authors used whole-brain grey matter densities from MRI scans with SVM as the classifier and concluded that SVM models trained with less than $130$ samples results in an unstable model. The key difference of the study \cite{Nieuwenhuis2012} from previous similar studies \cite{de2018identifying,guo2020support}, was utilizing a large dataset and using an entirely separate dataset to perform the validation. Additionally, noting that typical schizophrenia medications affect the striatum (part of the brain), they masked it out, ensuring the model doesn't relate medication effects to Schizophrenia detection.  
Each imaging technique provides a different view of the brain functioning. To get the benefit of different imaging techniques, a multimodal classification model \cite{10.3389/fnhum.2013.00235} combined $3$ different data types: resting state functional MRI (rs-fMRI), \color{black} Diffusion tensor imaging (DTI) and sMRI. While the idea proposed in \cite{10.3389/fnhum.2013.00235}, don't have the main focus on classification but to design and evaluate a multivariate method which can find cross-information in more than two data types. 
In \cite{JingSui2014}, multi-set canonical correlation analysis (MCCA) was used to  combine the  data of rs-fMRI,  Electroencephalogram (EEG) \color{black}and sMRI and  proposed ensemble feature selection approach which resulted in very high prediction performance approaching $100\%$ by utilizing the additional modalities. Though they also concluded that combining multiple modalities does not always result in an enhanced result. A similar study \cite{10.1002/hbm.24863} which used rs-fMRI and sMRI with a similar outcome of increased accuracy when compared to single modalities. It can easily be realized what the major downfall of this multi-modal training scheme is: lack of data. 
Some datasets combined from various sources reach the $250$ marks, like the one used in \cite{Nieuwenhuis2012} or in \cite{dePierrefeu2018}, but most datasets sit at $80$ samples.

 Gaining insight from the previous studies about the feasibility and reliability of individual classification based on the sMRI, a novel machine-learning algorithm provided an interpretable brain signature \cite{dePierrefeu2018}. Instead of behaving like a ``Black Box" spewing out predictions, the model provided insights into the neuroanatomic markers aiding clinical interpretability. This was achieved using ElasticNet Total Variation (Enet-TV) \cite{HadjSelem2018} penalty, which gave Structured sparsity (which is a sparse and structured pattern of predictors). Furthermore, they pitted the Enet-TV against other SVM algorithms, showing similar predictive performance across the board. In addition to providing a clinically interpretable model, their research also suggested a shared neuroanatomical signature for early or late-stage Schizophrenia patients. 
  The study in \cite{10.1002/hbm.24863}  attempted to identify the significantly contributing brain regions by averaging the weights across the five datasets used and reported the top $10$ brain regions.

Until now, the papers discussed have undertaken the entire brain, sometimes obfuscating certain regions to remove the effects of medications. But studies like \cite{LI201576} suggest that only particular brain regions, i.e. caudate nuclei, thalami and right side amygdala, are significant in identifying a Schizophrenia patient. 
 Since Schizophrenia patients have structural changes in hippocampus and amygdala regions, and the study \cite{guo2020support} extracted only the hippocampus and amygdala regions of the brain for the classification of Schizophrenia subjects and concluded that hippocampal and amygdaloid structures could be utilized for classification.
\vspace{-4mm}
\section{Subjects and Methods}
\label{Sec:Subjects and Methods}

\subsection{Subjects}

The data used in this study is obtained from the Center for Biomedical Research Excellence (COBRE) data set (Available at http://fcon\_1000.projects.nitrc.org/indi/retro/cobre.html). Data consists of $72$ subjects with schizophrenia ($38.1$ ± $13.9$ years old, range $18-65$ years) and $74$ age-matched healthy control ($35.8$ ± $11.5$ years old, range $18-65$ years).
\vspace{-4mm}
\subsection{3D MRI Processing}

Image processing was performed using the CAT12 package   
 (http://dbm.neuo.uni-jena.de) implemented in the Statistical Parametric Mapping (SPM) toolbox version 12 
 (http://www.fil. ion.ucl.ac.uk/spm/software/spm12/). In summary, 3D T1-weighted MRI scans were parcellated into grey matter (GM), white matter (WM) and cerebrospinal fluid, skull, scalp and air cavities.  In this study, the GM and WM tissues have been examined. Using a high-dimensional Diffeomorphic Anatomic Registration Through Exponentiated Lie algebra algorithm (DARTEL), the GM and WM images were normalized into Montreal Neurological Institute (MNI) space. The smoothed GM and WM images were generated through an $8$-mm full-width-half-maximum Gaussian kernel. The GM and WM images were visually inspected after each step in the preprocessing phase. Besides, we used the quality check procedure implemented in the CAT12 toolbox to identify possible outliers. Finally, we re-sampled GM and WM images to $4$-mm isotropic spatial resolution and extracted GM and WM voxel values from whole-brain data (i.e., total of $29,852$ voxel values per modality) as raw features for classification tasks. 
\vspace{-4mm}
\subsection{Classification Algorithms}
    The classification algorithms evaluated in this study for schizophrenia are explained below. Detailed information on the algorithms is available in the Supplementary file.
    \subsubsection{Random Forest (RaF) \cite{breiman2001random}}
    Random Forests proposed by Leo Breiman et al. in 2001 is a collection of tree predictors (also called tree-structured classifiers, which at their core are nested if-else statements used to vote for classes) where each tree is generated via independent and identically distributed random vectors. It has been shown that a sufficiently large forest always converge and a forest generated using random features generally produces better accuracy than a single tree classifier. RaF combined both the concepts of bagging and random subspace which improved its generalisation performance.
    \subsubsection{Oblique RaF (MPRaF-T, MPRaF-P and MPRaF-N) \cite{zhang2014oblique}}
    Oblique RaF was proposed to handle multiclass classification with an improved geometric property. {Multisurface Proximal Support Vector Machine (MPSVM)\cite{mangasarian2005multisurface}} is used to generate clustering hyperplanes at the non-terminal nodes of a decision tree. Now, RaF is implemented using the MPSVM-based decision trees and then subsequently using various regularisation methods. It was shown that Oblique RaF performs better than RaF and have significantly less variance and bias. MPRaF-T, MPRaF-P, and MPRaF-N represent the MPSVM-based RaFs with Tikhonov, axis-parallel, and NULL space regularization, respectively. 
    \subsubsection{Heterogeneous RaF \cite{katuwal2020heterogeneous}}
    As noted by \cite{zhang2014oblique}, RaF's data splitting leads to axis-parallel decision boundaries, which can lead to poor utilization of the geometric property of the data. But \cite{katuwal2020heterogeneous} noted that even though Oblique RaF allows for oblique splits, it is sub-optimal. The Heterogeneous RaF uses diverse linear-classifiers at the tree's nodes and searches for the best split at every node by optimizing the impurity criteria. Heterogeneous RaF Forests are shallower and faster to train than RaF. For the decision trees, each split is rated based on impurity criterion. All the splits at each non-leaf nodes are linked with an impurity measure. The one which is having the maximum value is the selected split for that particular node. The six different classifiers which have been employed are SVM, MPSVM, Linear Discriminant Analysis (LDA), Least Squares SVM (LSSVM), Ridge Regression (RR) and Logistic Regression (LR) as they have performed well in several domains \cite{fernandez2014we}. 
    \subsubsection{Kernel Ridge Regression (KRR) \cite{rakesh2017ensemble}}
    One of the kernel-based methods is the Kernel Ridge Regression (KRR). The KRR has a closed-form solution which lends it to faster training. Despite being relatively straightforward than other members of kernel-based methods such as SVM, it can produce comparable results. The kernel ridge regression method is based on Ridge Regression and Ordinary Least Squares. 
    \subsubsection{$K$ nearest neighbours (KNN) \cite{hand2007principles}}
    $K$ nearest neighbours algorithm assigns the label depending upon the similarity of the point with its neighbours. A constant $K$ is first chosen for the algorithm. The Euclidean distance of the given point is calculated and the $K$ nearest members are selected from it. The number of data points is counted and new data points are assigned to the category for which there are maximum number of neighbours. The number of nearest neighbours, $K$, in our case is $5$.
    \subsubsection{Neural Networks \cite{kingma2014adam}}
    Neural networks are network of node layers comprising of an input layer, multiple hidden layers and an output layer. Each layer has multiple number of nodes and the nodes of each layer are interconnected with the other layers. The output of each of the layer is calculated through an activation function and the output of activation layer of the last layer is the final output. {Adam optimization technique \cite{kingma2014adam}} has been used in order to tune the parameters. 
    \subsubsection{Random vector functional link network (RVFL) \cite{zhang2016comprehensive}} 
    RVFL is the randomized version of the functional link neural network. It shows that from the input layer to the hidden layer, the value of weights can be generated randomly in a suitable domain and fixed in the learning stage. The closed-form based RVFL obtains the output weights in a single-step and exhibits a higher efficiency than the iterative method.
    \subsubsection{Random vector functional link network with Auto Encoder (RVFLAE)\cite{zhang2019unsupervised}}
    Autoencoder is an unsupervised learning model for which the output and input layers share the same neurons in order to reconstruct its own inputs. In this method, we adopt a sparse autoencoder to learn appropriate network parameters of RVFL, which are developed via $l_1$norm optimization instead of the usual $l_2$ norm retaining more informative features. 
    \subsubsection{Support vector machine (SVM) \cite{cortes1995support}}
    SVM is a binary classification algorithm which classify the labelled data in two classes. SVM generates an optimal hyperplane using data to separate the classes. Since there may be more than one hyperplane possible for that, SVM finds the optimal hyperplane to do the classification by solving a Quadratic Programming Problem (QPP). Thus, a new data point can be classified based on the optimal hyperplane formed by SVM. And if the given data is not linearly separable, SVM do the task using kernel method.
    \subsubsection{Twin support vector machine (TWSVM) \cite{khemchandani2007twin}}
    Inspired from SVM, TWSVM is a classification algorithm which classifies the given labelled data into two classes by generating two non-parallel hyperplanes. TWSVM solves two smaller sized QPPs, unlike SVM, each for one class. The two required planes are formed by solving a problem which minimize the distance of points of corresponding class to the plane and keep it as far as possible from another class. Then, a new data point is assigned a class by calculating its distance from the planes.  
    \subsubsection{Twin bounded support vector machine (TBSVM) \cite{shao2011improvements}}
    TBSVM is an  improved version of TWSVM which modified the optimizations problem in TWSVM to give better performance and thus making the classification more accurate. It added an extra regularization term in the formulation of TWSVM which applied the structural risk minimization principle in the model. 
    \subsubsection{Least square twin support vector machine (LSTSVM) \cite{kumar2009least}}
    LSTSVM is the least squares version of TWSVM. The formulation of LSTSVM leads to fast and simple algorithm to generate the two non-parallel hyperplanes for binary classification. The two primal problems used to find the required hyperplanes are formulated in least squares sense i.e. using the equality constraints instead of the inequality constraints. The problem in LSTSVM can be solved very easily and simply by solving a system of two linear equations.   
    \subsubsection{Robust energy based least square twin support vector machine (RELSTSVM) \cite{tanveer2016robust}}
    Robust energy based LSTSVM, proposed by Tanveer et al., is another extension of TWSVM which adds a maximum margin regularization term in primal problem and moreover uses an energy parameter in the constraints, which helped in lessening the effect of noise in the data. According to a recent study \cite{tanveer2019comprehensive},  RELSTSVM model leads to better classification performance among the TWSVM models.
    \subsubsection{Pinball general twin support vector machine (PinGTSVM) \cite{tanveer2019general}}
    Pinball general twin support vector machine also generates non-parallel hyperplane for classification, similar to TWSVM, but uses pinball loss function in place of hinge loss without affecting the computational complexity of the algorithm. The use of pinball loss function makes it less sensitive to noise in classification of data and make it more stable for re-sampling of data.
    \vspace{0.1cm}
    
\subsection{Feature Selection Methods}
The feature selection methods evaluated in this study for schizophrenia are explained below:
\subsubsection{RankFeatures() function and its various criterions}
The \textit{rankfeatures()} is a MATLAB\textsuperscript{\textregistered} \cite{MATLAB:2021} function which ranks key features by class separability criterion. It uses various independent evaluation criteria to assess the significance of features. The criterion here refers to an objective function that minimises the overall feasible feature subset. The \textit{rankfeatures()} uses the following feature independent criteria:
\paragraph{T-test \cite{THEODORIDIS2009261}}
The ``ttest” is the \textit{default}, criteria used by the \textit{rankfeatures()} function. The T-test ranks the features based on an absolute value, two-sample T-test with pooled variance estimate. The ``ttest” criterion assumes that the classes are normally distributed.
\paragraph{Entropy \cite{kullback1997information}}
The ``entropy” criterion uses the relative entropy, also known as Kullback-Liebler distance or divergence. The ``entropy” criterion assumes that the classes are normally distributed.
\paragraph{Bhattacharyya \cite{THEODORIDIS2009261}}
The ``Bhattacharyya” criterion uses the minimum attainable classification error, or Chernoff bound to rank features. The ``Bhattacharyya” criterion assumes that the classes are normally distributed.
\paragraph{ROC \cite{THEODORIDIS2009261}}
The ``ROC” criterion uses the area between the empirical receiver operating characteristic (ROC) curve and the random classifier slope to rank features. The ``ROC” criterion is a non-parametric test.
\paragraph{Wilcoxon \cite{wilcoxon1992individual}}
The ``Wilcoxon” criterion uses the absolute value of the standardised U-statistic of a two-sample unpaired Wilcoxon test, also known as Mann-Whitney, to rank features. The ``Wilcoxon” criterion is also a non-parametric test.

\subsubsection{Minimum redundancy maximum relevance (MRMR) algorithm \cite{ding2005minimum}}
The MRMR algorithm is a sequential feature selection method that finds an optimal set of mutually and maximally dissimilar features. The MRMR performs this by maximising the relevance of the feature set to the response variable and minimising the redundancy of a feature set.

\subsubsection{Neighborhood Component Analysis (NCA) \cite{yang2012neighborhood}}
NCA is a non-parametric feature selection method used explicitly for regression and classification algorithms. The feature weights (importance of a feature) are obtained using a gradient ascent technique to maximise the expected leave-one-out classification accuracy with a regularisation term.

\subsection{Validation, Experimental Setup}
This study used Matlab\textsuperscript{\textregistered} R2021a\cite{MATLAB:2021} to implement all the required code for the different methods. The functions used were (but not limited to): \textit{rankfeatures()}, \textit{fscmrmr()}, \textit{fscnca()}. In all experiments, $10$-fold cross-validation was used. To study the variation of accuracy with increasing number of features and to obtain the minimum or the optimal number of features, we experimented with $100 - 1300$ (with a step size of $100$) selected features. The various hyper parameter ranges used for various methods have been tabulated in Table \ref{table:parameters}. The classification accuracies corresponding to different classification models versus feature selection approaches for the combined matter at 500 features are available in Table \ref{table:Acc_FS_combined_matter}. The results of 500 features are presented as maximum accuracy for Integrated GM and WM occurs at the same. The Tables corresponding to WM and GM are available in the Supplementary file as Table S-2 and S-3.

\begin{table}[ht!]
    \centering
    \tabcolsep=0.11cm
    \begin{tabular}{lll}
        \hline
        Parameter Name            & Symbol       & Range/Value                         \\
        \hline
        \hline
        Penalty parameters        & $c_i\footnotemark[1]$ & \{$10^i | i = -5:5$\} \\
        (for TSVM-based models)\\
        \hline
        Non-linear kernel parameter & $\gamma$        & \{$2^i|i = -10:10$\}      \\
        \hline
        NCA regularisation term    & $\lambda$       & \{$2^i/\text{N\footnotemark[2]}|i = 1:20$\}           \\
        \hline
        RELSTSVM parameter      & $E$   & \{$0.5, 0.6, 0.7, 0.8, 0.9, 1$\}                 \\
        \hline
        RVFL \& RVFL-AE        & $C$     & \{$-5:1:14$\}        \\
        parameters          & $N$ & \{$3:20:203$\} \\
        \hline
        Ensemble size of the trees & & $100$\\
        for RaF methods\\
        \hline
        pinGTSVM parameter & $\epsilon$ & $0.05$\\
        \hline
    \end{tabular}
    \caption{Parameter Ranges and values for various methods. ($^1i = 1,2,3,\dots$, $^2$Number of Samples)}
    \label{table:parameters}
    \vspace{-9mm}
\end{table}

\section{Results}
\label{Sec:Results}
The performance of the classification models varies with different feature extraction methods and the number of features selected. We discuss the performance of the models with GM, WM and the integration of GM and WM data\color{black}. When discussing classification models, accuracy means the average accuracy across both feature extraction techniques and the feature range, unless otherwise stated. Similarly, the accuracy assigned to a given feature extraction technique represent the average accuracy across the different classification models. Tables of other metrics i.e. AUC, Sensitivity, Specificity, Precision, F-Measure and G-Mean are available in the Supplementary file.

\subsection{White Matter}
Across the entire feature range (i.e. 100 - 1300), the TWSVM based classification models showed the highest average accuracy compared to the rest of the classification models, as can be seen in figure S-1a. The rest of the families, i.e. RaF, Neural Networks, KNN and KRR, stay together through the feature range. Among SVM-based classifiers, the non-linear RELSTSVM ($75.16\%$) and non-linear TBSVM ($74.30\%$) achieved the highest average accuracy. Non-linear TWSVM followed them with $73.24\%$ accuracy, closely followed by non-linear LSTSVM at $72.47\%$. The linear TBSVM and RELSTSVM showed ${\raise.17ex\hbox{$\scriptstyle\sim$}}70\%$ accuracy. The lowest-performing models are pinGTSVM and RVFLAE with ${\raise.17ex\hbox{$\scriptstyle\sim$}}60.7\%$ accuracy. Heterogeneous-RaF achieves the maximum accuracy of $84.04\%$ for WM with 900 features selected using Wilcoxon feature selection.

Among the RaF based models, Heterogeneous-RaF shows the best performance, with an average accuracy of $67.70\%$. In contrast, MPRaF-T shows the lowest average accuracy with $62.93\%$. Among the variants of neural networks, the standard neural network and randomized based neural network show ${\raise.17ex\hbox{$\scriptstyle\sim$}}65.5\%$ average accuracy. The autoencoder based RVFL model showed the lowest average accuracy ($60.89\%$) among the neural network models. Also, the non-linear kernel-based KRR model (accuracy $67.15\%$) is better than the linear kernel-based KRR model (accuracy $63.36\%$) in terms of average accuracy.

Discussing the feature selection methods we can refer to S-2a, the Wilcoxon is the best choice across the entire feature range with an average accuracy of $76.16\%$. The NCA, in addition to being the worse performing method with an average accuracy of $61.50\%$ it is also unstable with the number of features selected. ROC, Entropy and Bhattacharya all perform with an average accuracy of ${\raise.17ex\hbox{$\scriptstyle\sim$}}67.1\%$. MRMR and T-test follow with $65.42\%$ and $62.51\%$ average accuracy, respectively.

Observing that Wilcoxon feature selection performs better than all other methods for the entire feature range, comparing classifiers based on Wilcoxon feature selection is more beneficial than average. Since we have already discussed classifiers with respect to (w.r.t.) average, we now look at classifiers which significantly different. When comparing w.r.t. Wilcoxon, the Neural Network becomes the best classifier (averaged across all feature range) at $80.91\%$. Heterogeneous-RaF also performs significantly better, achieving $80.42\%$ at rank 4. Additionally, RaF-LDA also performs much better with $79.31\%$ and rank 6.

\subsection{Grey Matter}
The grey matter has some exciting results both in terms of classification techniques and feature selection methods. From Figure S-1b, it is obvious to see that the SVM family of classifiers perform better than the rest of the families for the entire feature range. Another observation can be made for the neural networks that they perform worse than all other families throughout the feature range. Non-linear TBSVM and non-linear RELSTSVM achieved the highest $73.36\%$ and $72.64\%$ average accuracy, respectively, followed by non-linear TWSVM ($72.03\%$) non-linear LSTSVM ($70.79\%$). The lowest-performing models are pinGTSVM and RVFLAE with ${\raise.17ex\hbox{$\scriptstyle\sim$}}59.5\%$ accuracy. Non-linear RELSTSVM achieves the maximum accuracy of $83.99\%$ for grey matter with 1200 features selected using Bhattacharyya feature selection.

Among the RaF based models, the standard RaF method shows the best performance, with an average accuracy of $67.09\%$. In contrast, MPRaF-N shows the lowest average accuracy with $64.27\%$. Among the variants of neural networks, the standard neural network and randomized based neural network show ${\raise.17ex\hbox{$\scriptstyle\sim$}}65\%$ average accuracy. The RVFLAE model showed the lowest average accuracy ($59.06\%$) among the neural network models. Also, the non-linear kernel-based KRR model (accuracy $69\%$) is better than the linear kernel-based KRR model (accuracy $63.67\%$) in terms of average accuracy.

Discussing the feature selection methods we can refer to S-2b, the Wilcoxon is the best choice across the entire feature range with an average accuracy of $75.21\%$. The NCA again is very unstable but performs very well with an average accuracy of $68.08\%$. Entropy and Bhattacharya both perform terribly at lower features (dipping as low as $52\%$) but approach the best performing Wilcoxon at higher features. MRMR performs better than both ROC and T-test at $67.5\%$. ROC and T-test achieve an average accuracy of $62.62\%$ and $63.32\%$, respectively.

Wilcoxon feature selection performs better than all other methods for the entire feature range. Thus, we can compare classifiers based on Wilcoxon feature selection. When comparing w.r.t. Wilcoxon, the linear RELSTSVM ($78.85\%$) perform just marginally lower than the non-linear variation ($79.13\%$). The neural network ($78.28\%$) performs significantly better than when using the average, but it is not at the top.

\subsection{Integrated GM and WM}
\begin{figure}[htbp]
\vspace{-4mm}
\centerline{\includegraphics[width=0.5\textwidth]{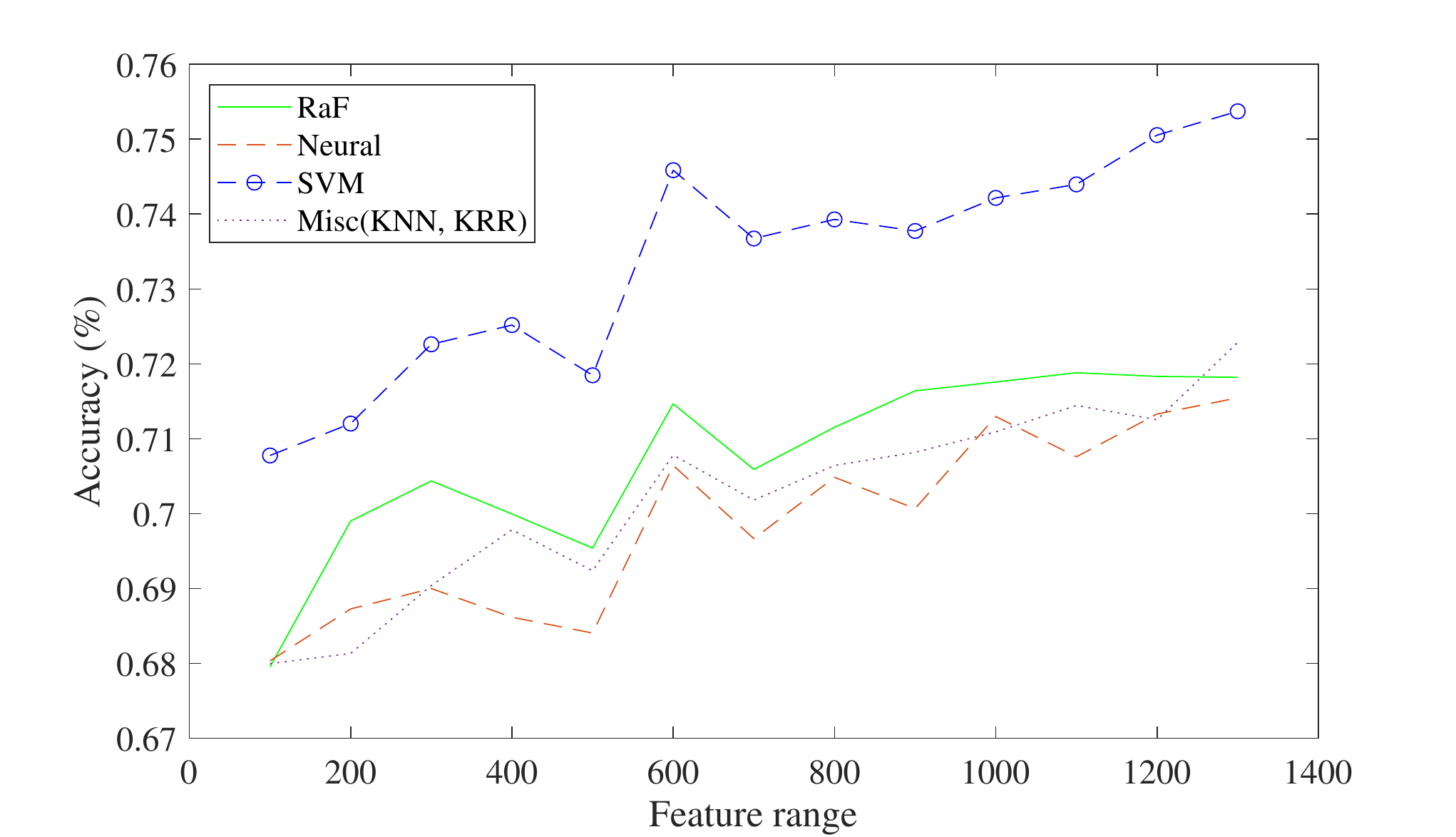}}
\caption{Average performance of classifiers families w.r.t Integrated WM and GM (Combined matter)}
\label{fig:CF_CM}
\vspace{-6mm}
\end{figure}

The combined matter (i.e. integrated GM and WM data) \color{black} achieves better results than individual grey matter and white matter. From Figure \ref{fig:CF_CM}, it can be inferred that the SVM based classifiers perform significantly better than the rest of the families for the entire feature range. The RaF based methods come next, followed by KNN and KRR. Neural networks again perform the worst. Non-linear TBSVM and non-linear RELSTSVM achieved the highest $78.47\%$ and $77.71\%$ average accuracy, respectively, followed by non-linear TWSVM ($77.62\%$) non-linear LSTSVM ($76.32\%$). The lowest-performing models are RVFLAE, pinGTSVM and KNN with ${\raise.17ex\hbox{$\scriptstyle\sim$}}66\%$ accuracy. The standard neural network achieves the maximum accuracy of $86.71\%$ for combined matter with 500 features selected using Wilcoxon feature selection.

Among the RaF based models, the Heterogeneous-RaF method shows the best performance, with an average accuracy of $73.31\%$. In contrast, MPRaF-T shows the lowest average accuracy with $68.92\%$. Among the variants of neural networks, the standard neural network and randomized based neural network show $73\%$ and $70.77\%$ average accuracy, respectively. The RVFLAE model showed the lowest average accuracy ($65.85\%$) among the neural network models. Also, the non-linear kernel-based KRR model (accuracy $72.13\%$) is better than the linear kernel-based KRR model (accuracy $71.15\%$) in terms of average accuracy.

\begin{figure}[htbp]
\centerline{\includegraphics[width=0.5\textwidth]{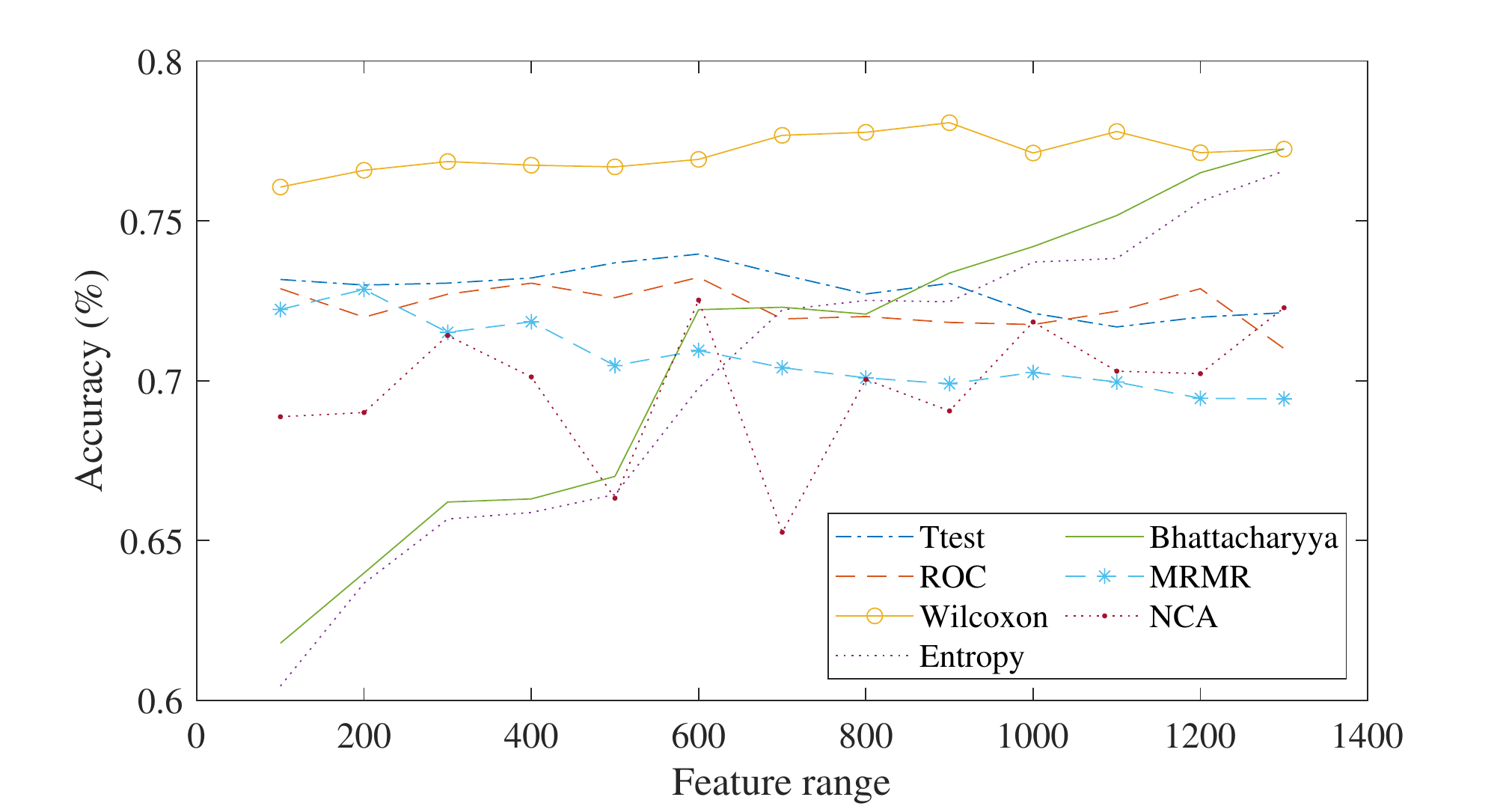}}
\vspace{-2mm}
\caption{Average performance of feature selection methods w.r.t Integrated WM and GM (Combined matter)}
\label{fig:FS_CM}
\vspace{-6mm}
\end{figure}

Discussing the feature selection methods we can refer to \ref{fig:FS_CM}, the Wilcoxon is the best choice across the entire feature range with an average accuracy of $77.12\%$. The NCA is unstable and performs poorly with an average accuracy of $69.78\%$. Entropy and Bhattacharya both start out being the worst performing methods at lower features (dipping as low as $60\%$) but approach the best performing Wilcoxon at higher features. ROC and T-test perform better than MRMR, which is at $70.71\%$. ROC and T-test achieve an average accuracy of ${\raise.17ex\hbox{$\scriptstyle\sim$}}72\%$.

Wilcoxon feature selection performs better than all other methods for the entire feature range. Thus, we can compare classifiers based on Wilcoxon feature selection. When comparing w.r.t. Wilcoxon, the standard neural network comes out on top ($83.98\%$), becoming the best classifier. The Heterogeneous-RaF becomes the second-best classifier at $82.36\%$. The linear TBSVM ($80.61\%$) and linear RELSTSVM ($79.75\%$) performs slightly better than non-linear TBSVM ($80.56\%$) and non-linear RELSTSVM ($79.62\%$).

\begin{figure}[htbp]
\centerline{\includegraphics[width=0.5\textwidth]{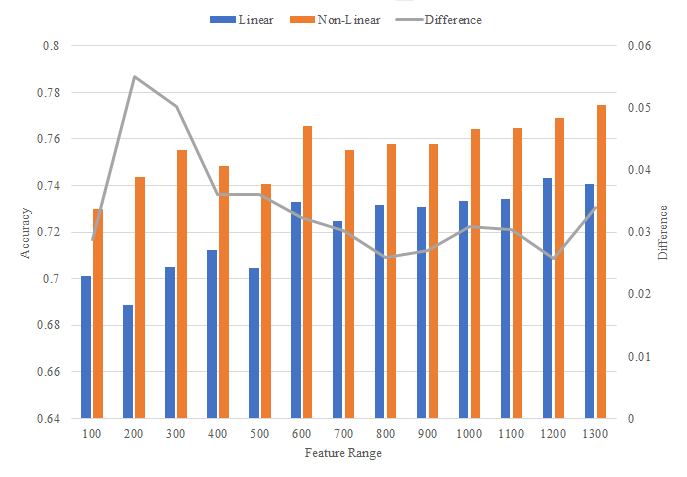}}
\vspace{-4mm}
\caption{Performance of linear and non-linear kernels for combined matter}
\label{fig:Lin_vsNL}
\vspace{-6mm}
\end{figure}

\begin{figure}[htbp]
\centerline{\includegraphics[width=0.5\textwidth]{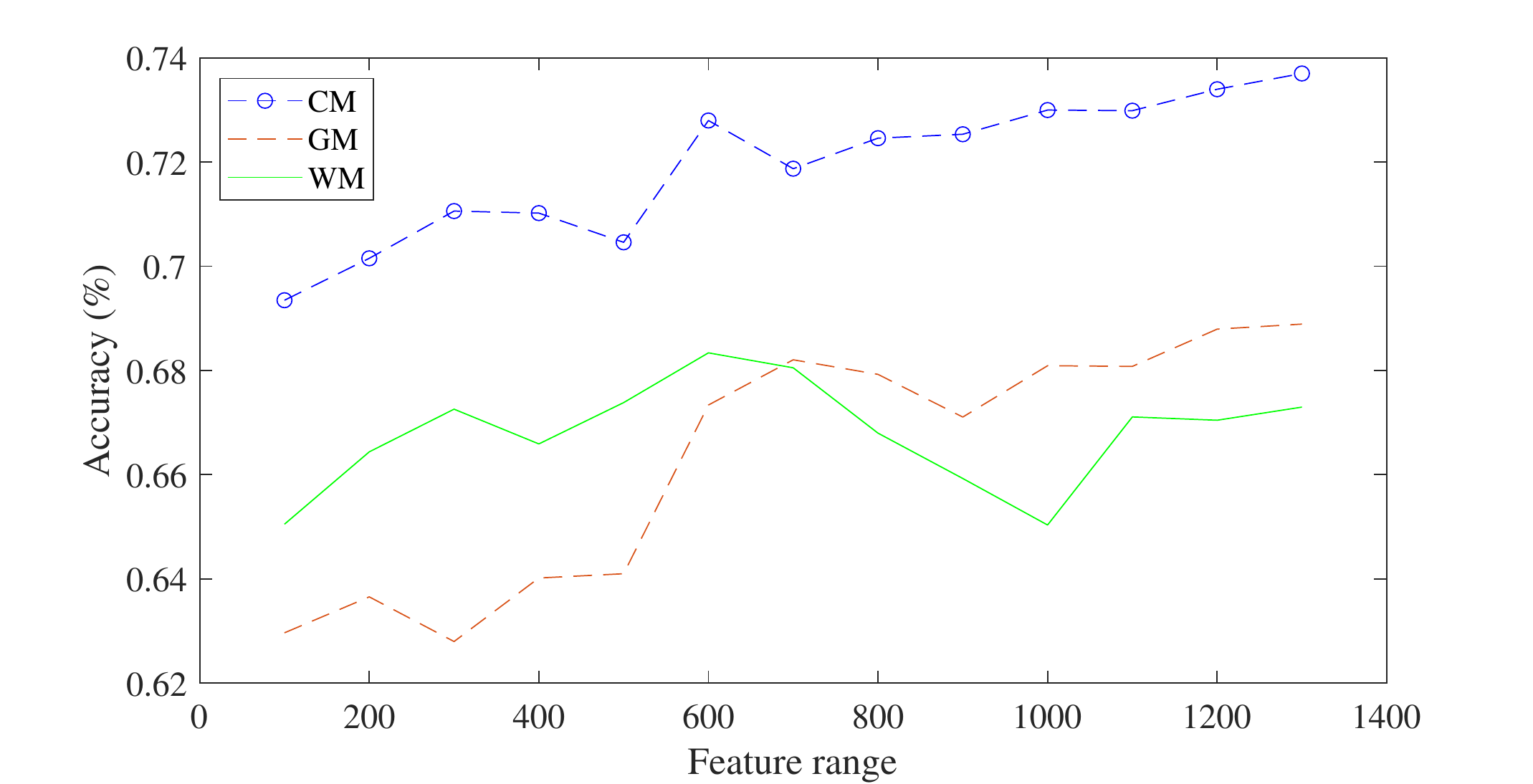}}
\caption{Average performance of classification algorithms with different matter types (i.e. WM, GM and Integrated GM and WM (CM)}
\label{fig:comparision_avg}
\vspace{-4mm}
\end{figure}

\begin{table}[htbp]
  \centering
  \tabcolsep=0.11cm
\begin{tabular}{llllllll}
 & \multicolumn{7}{c}{Feature selection methods}\\
\cline{2-8}
Classification  &T-Test &ROC &Wilc- &Entr- &Bhatta- &MRMR &NCA\\
techniques & & &oxon &opy &charyya & &\\
\hline
\multicolumn{8}{l}{\textbf{Linear variations:}}\\
\hline
\begin{filecontents*}{CM_500_results_lin.csv}
aa,aa,aa,aa,aa,aa,aa,aa
Het-RaF,72.57,73.95,82.1,67.95,62.29,74.62,63.62
KNN,69.95,71.81,73.9,60.48,60.48,65,69.67
MPRaF-N,73.1,71.67,72.48,64.33,62.95,69,61.62
MPRaF-P,71.71,75.19,75.9,61.57,67.19,70.33,69
MPRaF-T,74.52,75.86,75.1,55.29,53.33,68.29,69.1
Neural,77.38,78.76,86.71,61.57,64.24,72.29,66.52
pinGTSVM,68.57,67.76,75.1,58.95,58.86,58.1,64.14
RaF-LDA,75.33,73.29,81.38,54.62,64.33,69.71,65.81
RaF-PCA,74.57,74.48,71.62,66.29,68.33,69.81,69.9
RaF,71.67,71,75.14,64.48,65.05,75,71.05
RVFLAE,62.48,58.86,71.1,60.57,62.71,66.48,58.1
RVFL,73.33,72.76,76.48,71.19,65.57,68.48,60.95
KRR,73.14,71.86,75.24,62.76,64.1,74.43,67.1
LSTWSVM,71.85,65.62,75.4,69.28,68.25,67.33,63.71
RELSTSVM,74.28,73.18,78.84,69.12,70.7,72.14,67.8
SVM,77.43,71.95,71.71,66.24,66.9,67.1,61.71
TBSVM,75.52,73.9,80,68.29,69.05,72.95,67.86
TWSVM,73.48,67.81,78.57,68.29,68.29,69.86,66.48
\end{filecontents*}
\csvreader[late after line=\\,
            late after last line=,
            before reading={\catcode`\#=12},
            after reading={\catcode`\#=6}]
            {CM_500_results_lin.csv}
            {1=\methods,2=\ttest,3=\roc,4=\wilcoxon,5=\entropy,6=\bhattacharyya,7=\MRMR,8=\NCA}
            {\methods & \ttest & \roc & \wilcoxon & \entropy & \bhattacharyya & \MRMR & \NCA}\\
\hline
\multicolumn{8}{l}{\textbf{Non-linear (Gaussian Kernel)} variations:}\\
\hline
\begin{filecontents*}{CM_500_results_nl.csv}
aa,aa,aa,aa,aa,aa,aa,aa
KRR,73.19,73.9,72.48,66.43,67.14,73.19,67.76
LSTWSVM,74.6,74.52,79.84,78.38,76.9,73.26,69.75
RELSTSVM,76.21,76.32,79.83,79.37,79.7,76.56,69.75
SVM,77.48,71.95,71.71,61.48,64.81,66.29,61
TBSVM,78.19,78,79.95,79.52,78.86,76.62,69.76
TWSVM,77.95,77.95,79.95,78.1,78.1,74.52,69.62
\end{filecontents*}
\csvreader[late after line=\\,
            late after last line=,
            before reading={\catcode`\#=12},
            after reading={\catcode`\#=6}]
            {CM_500_results_nl.csv}
            {1=\methods,2=\ttest,3=\roc,4=\wilcoxon,5=\entropy,6=\bhattacharyya,7=\MRMR,8=\NCA}
            {\methods & \ttest & \roc & \wilcoxon & \entropy & \bhattacharyya & \MRMR & \NCA}\\
\hline
\end{tabular}
\caption{Accuracies for Integrated GM and WM for 500 features.}
\label{table:Acc_FS_combined_matter}
\vspace{-8mm}
\end{table}

\section{Discussion}
\label{Sec:Discussion}
This study aimed to perform a comprehensive evaluation of modern classification techniques and feature selection methods for schizophrenia classification. We have presented a basic overview of the different classification methods and evaluated them against different feature selection approaches on the same dataset. Selection of the optimal features and using the best available classification technique is essential for an MRI-based machine learning system aimed at early diagnosis. Our findings indicate that using twin SVM-based methods such as RELSTSVM or TBSVM performs best for nearly all matter types. The random forest based methods generally perform mediocrely. The worst performing classification models are the pinGTSVM, RVFL-AE, KRR, KNN and MPRaF-T.

Figure \ref{fig:Lin_vsNL} is constructed by averaging only the methods that have both linear and non-linear variants (i.e. SVM, KRR, TWSVM, TBSVM, LSTWSVM and RELSTSVM). Inferring from figure \ref{fig:Lin_vsNL}, a critical observation can be made for the performance of non-linear kernel functions against linear ones. In nearly all studied methods (pinGTSVM and standard SVM being the exception), the non-linear kernel function performs better than the linear variation. This observation follows suit with the result that a linear kernel is the degenerate version of the non-linear (RBF or Gaussian) kernel \cite{keerthi2003asymptotic}. Thus an adequately tuned non-linear kernel consistently out-performs the linear kernel. But, an observation can be made that with an increasing number of features, the advantage of using a non-linear kernel diminishes. This diminished performance can be attributed to the fact that at a higher number of features, one may not need to map features to a higher dimension \cite{hsu2003practical}. Thus, the time penalty to tune the kernel function (in case of non-linear) becomes outweighed by the rapid computation of the linear kernel when both kernels provide relatively similar accuracy. Thus, with a large number of features, one is better off using a linear kernel.

The feature selection methods significantly impact the performance of classification models, as one might expect. The Wilcoxon was the all-around best feature selection method, performing the best for all the different matter types and across the entire feature range. This observation is supported by previous studies \cite{tian2021evaluation} and \cite{liao2006gene}. The exceptional performance of Wilcoxon is especially prevalent when sample sizes are small or the data doesn't resemble Normal distribution. Entropy and Bhattacharya are fascinating methods. At lower feature numbers, they perform equally terribly, but at a higher number of features, they approach the best performance. This behaviour can be seen in figure S-2, especially in integrated and grey matter. The MRMR, ROC and T-Test all perform mediocrely, varying based on what matter type is used. At present, using our specific dataset, our findings indicate that we can classify schizophrenia patients with a maximum of $86.71\%$ accuracy when using a standard neural network with $500$ features from combined matter, selected using Wilcoxon. The main advantages of twin SVM based models like TWSVM, TBSVM, RELSTSVM, LSTSVM etc. over standard SVM is that they give competitive performance in terms of accuracy and reduce the computational complexity of SVM because these models generate two non-parallel hyperplanes instead of one single hyperplane in SVM which leads to solving two smaller sized Quadratic Programming Problems (QPPs) instead of one larger QPP in SVM. The paper \cite{tanveer2019comprehensive} concludes that twin SVM based models performs better than other family of classifiers.

Referring to figure \ref{fig:comparision_avg}, our results strongly suggest that using both grey matter (GM) and white matter (WM), i.e. integrated matter\color{black}, leads to improved performance for the classification of schizophrenia patients. The GM performs better than WM after reaching a threshold number of features (in our case $700$), but the results shoot up by a substantial margin (${\raise.17ex\hbox{$\scriptstyle\sim$}}4\%$) when the combined matter is used. It can be seen from figure \ref{fig:comparision_avg} that this is the case for all the evaluated classification techniques and that there are no exceptions for this observation.

\vspace{2mm}

\section{Conclusions}
\label{Sec:Conclusions}
In this study, we comprehensively evaluated various classification models to identify the best available machine learning model for the classification of schizophrenia subjects. We assessed $25$ classification models involving the variants of support vector machines, twin support vector machines, random forest, Kernel ridge regression and neural networks. Additionally, we evaluated $7$ feature selection methods: Wilcoxon, MRMR, ROC, Entropy, T-Test, Bhattacharyya and NCA.  Moreover, these evaluations were conducted on the features based on grey matter (GM)\color{black}, white matter (WM) \color{black} and the integrated GM and WM data\color{black}.

The contributions from this paper are four-fold. First, we underlined how different families of machine learning algorithms perform with the schizophrenia dataset. We found that, for the most part, the Non-linear twin SVM-based family of classifiers outperform all other classifiers. This family includes (in the order of gradually worsening performance) RELSTSVM, TBSVM, TWSVM and LSTSVM. However, the pinGTSVM is the worst-performing family member, ranking the lowest across all the classifiers. On the other hand, the KNN, KRR, MPRaF-T and non-linear SVM are the lowest performers. Most RaF-based methods occupy the middle of the spectrum. An additional observation is that, for the most part, the non-linear variant of a method outperforms the linear variation.

The second contribution is that we evaluated the performance of different feature selection methods. Our results indicate that Wilcoxon is the best performing methods with a top rank across all the matter types. Entropy and Bhattacharyya have improved performance with an increasing number of features. NCA is an unstable method, although it had a good average performance. T-Test, ROC, MRMR are also reasonable choices for feature selection, but they are not recommended.

Third, we found that utilising both grey and white matter for classification yields better results than any individual matter type. 

In conclusion, the feature selection method, the number of features selected and the classification model should be appropriately chosen for better generalisation performance on the classification of schizophrenia subjects. This study recommends using standard neural network, RELSTSVM, TWSVM, or heterogeneous-RaF as the classification model with ${\raise.17ex\hbox{$\scriptstyle\sim$}}700-1200$ features selected via Wilcoxon feature selection method for better generalisation performance on the classification of schizophrenia datasets. We hope that the evaluation presented in this paper encourages future research to use better classification algorithms and feature selection algorithms for clinical dataset classification.

New developments in machine learning are rapid and can improve the results of previous algorithms by a significant margin. In the future, much scope remains for the development of better specific models. Therefore, these new variants or methods need to be tested on real-life datasets such as schizophrenia to grasp their viability. In the future, one can extend this study by various margins, i.e. (1) the dataset can be enhanced to utilise data from multiple sources; thus, it should be evaluated if combining data from various sources (i.e. MRI images with varying scanning parameters such as slice thickness, field of view, bandwidth, repetition time, etc.) leads to better generalisation or if it results in a worse performance. (2) This study utilised a single feature extraction technique (i.e. DARTEL); thus, future research for the effect of different feature extraction techniques needs to be conducted.  (3) This study used a single modality (i.e. sMRI) and usage of different modalities  including the Functional MRI (fMRI),  Electroencephalogram (EEG) should also be evaluated using a similar setup in future studies. \color{black}
 The source code will be available at \underline{https://github.com/mtanveer1}
\vspace{0.2cm}

\section*{Acknowledgment}
This work is supported by Science and Engineering Research Board (SERB), Government of India under Ramanujan Fellowship Scheme, Grant No. SB/S2/RJN-001/2016, and Council of Scientific \& Industrial Research (CSIR),
New Delhi, INDIA for funding under Extra Mural Research (EMR)
Scheme grant no. 22(0751)/17/EMR-II. We gratefully acknowledge the Indian Institute of Technology Indore for providing the required facilities and support for this work.

\bibliographystyle{model1b-num-names}
\bibliography{refs}

\end{document}


\startsupplement
\title{Supplementary file of Diagnosis of Schizophrenia: A comprehensive evaluation}
\author{}
\maketitle

\section{Classification Algorithms}
The classification algorithms evaluated in this study for diagnosis of schizophrenia are explained below.
Let the training set be  $((x_1, y_1),\cdots,(x_N, y_N))$ where $N$ is the total number of training examples. X is the feature matrix [$x_1,x_2,\cdots,x_N$] of size $N\times d$ and Y = [$1,2,\cdots,m$] is a $N\times 1$ vector of class labels.
\vspace{-4mm}
    

\subsection{Support Vector Machine (SVM)}
Support Vector Machine \cite{cortes1995support} is supervised learning algorithm which classifies the given data into two classes by constructing a hyperplane. From the many possible hyperplanes, SVM chooses the one which maximize the margin between the two classes of data points.\\
SVM finds the optimal hyperplane by solving the following optimization problem,
\begin{align}
    \underset{w,\xi }{min}&~~\frac{1}{2}||w||^2 + C\sum_{i=1}^m\xi_i,\nonumber\\
    s.t.&~~ y_i(w^Tx_i+b) \geq 1-\xi_i\nonumber\\
    &\xi_i \geq 0,~~ i=1,2,\dots,m
\end{align}
where $w,x \in \mathbb{R}^n$, $\xi$ is the degree of misclassification and C is the penalty parameter. Using Karush-Kuhn Tucker (KKT) conditions \cite{fletcher2013practical}, we can solve the above problem.\\
The optimal hyperplane is given as $w^Tx+b=0$. \\
SVM underperforms in places where number of features for each data point exceeds the number of training data samples.
\vspace{-4mm}
\subsection{Twin Support Vector Machine (TWSVM)}
TWSVM \cite{khemchandani2007twin} divides the given data into two classes by generating two non-parallel hyperplanes by minimizing the distance of each class points from its corresponding hyperplane. So, TWSVM solves a pair of quadratic programming problems for two classes of data points.\\
The pair of QPPs is given as:

 \begin{align}
        \underset{w_1,b_1,\xi }{min}&~~\frac{1}{2}||X_1w_1+e_1b_1||^2 + c_1e_2^T\xi\nonumber\\
        s.t.&~~  -(X_2w_1+e_2b_1) \geq e_2-\xi,~~ \xi \geq 0 
 \end{align} and
\begin{align}
        \underset{w_2,b_2,\eta }{min}&~~\frac{1}{2}||X_2w_2+e_2b_2||^2 + c_2e_1^T\eta\nonumber\\
        s.t.&~~ (X_1w_2+e_1b_2) \geq e_1-\eta,~~ \eta \geq 0,
\end{align}

where $c_1,c_2$ are penalty parameters and $\xi,\eta$ are slack variables. Solving the above primal problem by taking Lagrangian and then using K.K.T. conditions \cite{fletcher2013practical}, we get the dual of the respective problems as follows:

\begin{align}
    \underset{\alpha}{max}&~~e_2^T\alpha-\frac{1}{2}\alpha^TG(H^TH)^{-1}G^T\alpha\nonumber\\
    s.t.&~~ 0 \leq \alpha \leq c_1
\end{align} and
\begin{align}
    \underset{\gamma}{max}&~~e_1^T\gamma-\frac{1}{2}\gamma^TP(Q^TQ)^{-1}P^T\gamma\nonumber\\
    s.t.&~~ 0 \leq \gamma \leq c_2
\end{align}
where $H=[X_1~~ e_1], G=[X_2~~ e_2], P=[X_1~~ e_1], Q=[X_2~~ e_2]$
which can be solved to give $[w_i~~b_i], i=1,2$. The new data point $x_j$ can be assigned class i by the equation $class(x_j) = arg \min\limits_{i=1,2}|w_i^Tx_j+b_i|$

For non-linearly separable data points, appropriate kernel can be used to project data points to a higher dimensional space where they can be linearly separable.
\vspace{-4mm}
\subsection{Twin Bounded Support Vector Machine (TBSVM)}
TBSVM \cite{shao2011improvements} is an extension of TWSVM that introduces a maximum margin regularization term which improves the classification accuracy. The minimization of extra regularization term leads to maximizing the margin between the decision hyperplane and its parallel plane.\\
So, the formulation of TBSVM is given as follows:
\begin{align}
    \underset{w_1,b_1,\xi }{min}&~~\frac{1}{2}||X_1w_1+e_1b_1||^2 + c_1e_2^T\xi +\frac{1}{2}c_2(||w_1||^2+b_1^2)\nonumber\\
        s.t.&~~  -(X_2w_1+e_2b_1) \geq e_2-\xi,~~ \xi \geq 0 
\end{align} and
\begin{align}
      \underset{w_2,b_2,\eta }{min}&~~\frac{1}{2}||X_2w_2+e_2b_2||^2 + c_3e_1^T\eta + \frac{1}{2}c_4(||w_2||^2+b_2^2)\nonumber\\
        s.t.&~~ (X_1w_2+e_1b_2) \geq e_1-\eta,~~ \eta \geq 0,
\end{align}
where $c_1,c_2,c_3,c_4$ are the penalty parameters and $\frac{1}{2}c_2(||w_1||^2+b_1^2)$, $\frac{1}{2}c_4(||w_2||^2+b_2^2)$ are the extra regularization terms. Now, considering the Lagrangian and using K.K.T. conditions, one can get the dual of the above problems as:
\begin{align}
    \underset{\alpha}{max}&~~e_2^T\alpha-\frac{1}{2}\alpha^TG(H^TH+c_2I)^{-1}G^T\alpha\nonumber\\
    s.t.& ~~ 0 \leq \alpha \leq c_1
\end{align}
and for the second problem dual is
\begin{align}
    \underset{\gamma}{max}&~~e_1^T\gamma-\frac{1}{2}\gamma^TP(Q^TQ+c_4I)^{-1}P^T\gamma\nonumber\\
    s.t.& ~~ 0 \leq \gamma \leq c_3,
\end{align}
where $H=[X_1~~ e_1], G=[X_2~~ e_2], P=[X_1~~ e_1], Q=[X_2~~ e_2]$ and $I$ is the identity matrix of appropriate dimension. Solving the above dual problems one can get the required decision hyperplanes.
\vspace{-4mm}
\subsection{Least Squares Twin SVM (LSTSVM)}
LSTSVM\cite{kumar2009least}, proposed by Kumar et. al., is an extension of TWSVM. The formulation of LSTSVM leads to very simple and fast algorithm for generating the two non-parallel hyperplanes to classify the data points. LSTSVM formulation solves a system of linear equations instead of two QPPs.\\
So, the formulation of LSTSVM is given as: 
\begin{align}
        \underset{w_1,b_1,}{min}&~~\frac{1}{2}||X_1w_1+e_1b_1||^2 + \frac{c_1}{2}\xi^T\xi\nonumber\\
        s.t.& ~~ -(X_2w_1+e_2b_1) = e_2-\xi,   
\end{align} and
\begin{align}
        \underset{w_2,b_2}{min}&~~\frac{1}{2}||X_2w_2+e_2b_2||^2 + \frac{c_2}{2}\delta^T\delta\nonumber\\
        s.t.& ~~ (X_1w_2+e_1b_2) = e_1-\delta,  
\end{align}
where $c_1,c_2$ are penalty parameters and $\xi,\eta$ are slack variables. e is the vector of ones of appropriate dimension. By substituting the constraints in objective function for each problem, above equations can be solved to give a system of two linear equations as:
\begin{align}
    \begin{bmatrix}
    w_1\\
    b_1
    \end{bmatrix}=-(G^TG+\frac{1}{c_1}H^TH)^{-1}G^Te_2\\
    \begin{bmatrix}
    w_2\\
    b_2
    \end{bmatrix}=(H^TH+\frac{1}{c_2}G^TG)^{-1}H^Te_1
\end{align}
where $H=[X_1~~e_1]$ and $G=[X_2~~e_2]$, So, the required non-parallel separating hyperplanes can be generated by solving the above system of equations.
\vspace{-4mm}
\subsection{Robust Energy-based Least Squares Twin Support Vector Machine (RELSTSVM)}
RELSTSVM\cite{tanveer2016robust} is another extension of TWSVM which adds an extra regularization term to the objective function of each problem and incorporates the structural risk minimization principle. Moreover, RELSTSVM adds an energy to each hyperplane to lessen the impact of noise and outliers, thus making the algorithm more efficient.\\
Similar to LSTSVM, the constraints of RELSTSVM makes the distance of hyperplanes to the data points to be exactly 1. and the extra regularization term in objective function leads to the following formulation:
\begin{align}
    \underset{w_1,b_1}{min}~~&\frac{1}{2}||X_1w_1+e_1b_1||^2+\frac{c_1}{2}\xi^T\xi+\frac{c_2}{2}(||w_1||^2+b_1^2)\nonumber\\
     s.t.&~~  -(X_2w_1+e_2b_1) = E_1-\xi
\end{align}
and
\begin{align}
    \underset{w_2,b_2}{min}~~&\frac{1}{2}||X_2w_2+e_2b_2||^2+\frac{c_3}{2}\eta^T\eta+\frac{c_4}{2}(||w_2||^2+b_2^2)\nonumber\\
     s.t.&~~  (X_1w_2+e_1b_2) = E_2-\eta
\end{align}
where $c_1,c_2$, $c_3,c_4$ are positive parameters and $E_1,E_2$ are the energy parameters. Without help of any external toolbox, the above primal can be directly solved by substituting the constraint in the objective function and setting the gradient of the obtained function with respect to (w.r.t.) to $w_1$, $b_1$ equal to zero, we get the system of linear equations as
\begin{align}
    v_1 = -(c_1Q^TQ+P^TP+c_2I)^{-1}c_1Q^TE_1
\end{align}
and for the second QPP,
\begin{align}
    v_2 = (c_3P^TP+Q^TQ+c_4I)^{-1}c_3P^TE_2
\end{align}
where $v_i=\begin{bmatrix}
w_i\\
b_i
\end{bmatrix}$ for $i=1,2$, $P=[X_1~~e]$, $Q=[X_2~~e_2]$
 \vspace{-4mm}
\subsection{Pin-GTSVM}
Pin-GTSVM \cite{tanveer2019general} is an extension of TWSVM that uses pinball loss instead of hinge loss which helps in making the algorithm insensitive to noise in data and more stable for re-sampling of data. Pinball loss also puts penalty on the correctly classified points. Hence, Pin-GTSVM obtains pair of planes for classification of data points by solving the following pair of QPPs:
 \begin{align}
        \underset{w_1,b_1,\xi}{min}&~~\frac{1}{2}||X_1w_1+e_1b_1||^2 + c_1e_2^T\xi\nonumber\\
        s.t.~~&  -(X_2w_1+e_2b_1) \geq e_2-\xi,\nonumber \\
        & -(X_2w_1+e_2b_1) \leq e_2+\frac{\xi}{\tau_2}
 \end{align} and
\begin{align}
        \underset{w_2,b_2,\eta}{min}&~~\frac{1}{2}||X_2w_2+e_2b_2||^2 + c_2e_1^T\eta\nonumber\\
        s.t.~~& (X_1w_2+e_1b_2) \geq e_1-\eta,\nonumber\\
        &(X_1w_2+e_1b_2) \leq e_1+\frac{\eta}{\tau_1},
\end{align}
where $\tau_1,\tau_2 \in [0,1]$ are pinball loss parameters. Forming the Lagrangian and using the K.K.T. conditions, we can obtain the dual as:
\begin{align}
    \underset{\alpha-\beta}{max}&~~e_2^T(\alpha-\beta)-\frac{1}{2}(\alpha-\beta)^TG(H^TH)^{-1}G^T(\alpha-\beta)\nonumber\\
    s.t.& -\tau_2c_1e_2 \leq (\alpha-\beta)
\end{align} and
\begin{align}
    \underset{\gamma-\delta}{max}&~~e_1^T(\gamma-\delta)-\frac{1}{2}(\gamma-\delta)^TH(G^TG)^{-1}H^T(\gamma-\delta)\nonumber\\
    s.t.& -\tau_1c_2e_1 \leq (\gamma-\delta)
\end{align}
where $H=[X_1~~e_1],G=[X_2~~e_2]$ Solving the above dual one can get the two decision hyperplanes and a new data point $x\in\mathbb{R}^n$ can be assign a label for it's class using the equations of hyperplanes.
\vspace{-4mm}
\subsection{Random Forest}
A random forest (RaF) is an ensemble of axis-parallel decision trees that are trained independently. Decision trees in a random forest employ recursive partitioning of the training data into smaller subsets that further aid in classification by optimizing an impurity criterion such as information gain or gini index{\cite{criminisi2012decision}}. In RaF, each non-leaf node is associated with a split function $f(x,\Theta)$ where
\begin{align}
    f(x,\Theta) = 1 ;~~x(\Theta_1) < \Theta\\*0;~~otherwise
\end{align}
where $\Theta_1 ~\epsilon ~\{1,2,\cdots,d\}$ is the selected feature and $ \Theta_2 ~\epsilon ~ \mathbb{R} $  is a threshold. The outcome determines the child node where $\boldsymbol{x}$ is routed to. The leaf nodes of the tree can either store class probability distributions or class labels based on the training samples they receive. At the time of testing, a test sample ${x}$, each tree returns probability distribution $ p_t(y|{x}) $ and the label of the class is obtained as average or majority vote.
\begin{align}
    y^* ({x}) = arg~\underset{y}{max}\frac{1}{T} \sum_{t=1}^{T} p_t(y|{x})
\end{align}
Here T is the number of trees in the forest.\\
Random Forest requires high computational power and time as it combines numerous decision trees in order to determine the class.

\subsection{Oblique Random Forest}
Oblique RaF \cite{zhang2014oblique} was proposed to handle classification with an improved geometric property. {Multisurface Proximal Support Vector Machine (MPSVM) \cite{mangasarian2005multisurface}} is used to generate clustering hyperplanes in decision trees. Now, RaF is implemented using the MPSVM-based decision trees and then subsequently using various regularisation methods. It was shown that Oblique RaF performs better than RaF and have significantly less variance and bias. MPRaF-T, MPRaF-P, and MPRaF-N represent the MPSVM-based RaFs with Tikhonov, axis-parallel, and NULL space regularization, respectively.

\subsection{Heterogeneous Random Forest}
Even though some of the oblique random forest based \cite{zhang2014oblique} on linear classifier perform better consistently, they aren't always the best variant of oblique random forest for every dataset. Heterogeneous RaF \cite{katuwal2020heterogeneous} uses several linear classifiers for generating the separating of hyperplanes. Even when some of the linear classifier based variants have lower ranks, they can still be integrated forming a heterogeneous linear classifier based oblique random forest. This would require us to evaluate $n$ classifiers in $K$ binary partitions, hence requiring $nK$ number of evaluations at each node. They employed a hyper class based partitioning with one-vs-all partitioning using multiple linear classifiers at each node.

The six different classifiers which have been employed are Support Vector Machines (SVM), Multisurface Proximal SVM (MPSVM), Linear Discriminant Analysis (LDA), Least Squares SVM (LSSVM), Ridge Regression (RR) and Logistic Regression (LR) as they have performed well in several domains\cite{fernandez2014we}.

For the decision trees, each split is rated based on impurity criterion. All the splits at each non-leaf nodes are linked with an impurity measure Gini Index. The one which is having the maximum value is the selected split for that particular node. Instead of looking for optimal oblique split in whole search space, the recursive partitioning property exhibited by decision trees, generating few oblique splits and used their $g(i)$ for selecting the best oblique split.

The ideal gini score $(g_i)$ is the one which is obtained when all the samples of one class are perfectly seperated from other class by an oblique split. By training the linear classifiers on partitions with higher $g_i$ and are likely to give higher $g$. One can ignore the partitions with lower $g_i$ which are likely to give lower $g$.

\begin{align}
    \underset{i=1,...,k}{min} \Bigg[\underset{i=i+1,...,k}{min}\left( \frac{diss(c_i,c_j)}{max_{m=1,...,k}diam(c_m)} \right) \Bigg]
\end{align}
Here $diss(c_i,c_j)= min_{x\epsilon c_i, y\epsilon c_j} ||x-y||$  is the dissimilarity between cluster $c_i$ and $c_j$ and $diam(c)=max_{x,y\epsilon C}||x-y||$ is the intra cluster function. We use Bhattacharyya distance as the metric for distance.

\subsection{Kernel Ridge Regression}
Kernel Ridge Regression \cite{rakesh2017ensemble} and SVM \cite{cortes1995support} are the best known members using kernel method. Kernel based methods are very useful when there is non-linear structure in data. KRR is faster to train and simpler with its closed form solution and can achieve performance which is comparable to complex methods such as SVM.

The kernel ridge regression method is based on {Ridge Regression \cite{saunders1998ridge}} and Ordinary Least Squares. The OLS minimizes the loss $\underset{\beta}{min}||Y-X\beta||^2$ which is the $L_2$ norm. A shrinkage parameter $\lambda$ is added to control the trade-off between variance and bias in the above expression giving us the following problem.
\begin{align}
    \underset{\beta}{min}||Y-X\beta||^2 + \lambda ||\beta||^2
\end{align}
The closed form solution for above can be problem given as $\beta = (X^TX + \lambda I)^{-1}X^TY $.  The label predicted for the new unlabeled example $x$ is given as $\beta^Tx$. The Kernel ridge regression method extends linear regression into non-linear and high-dimensional space. The data which is present in $X$ is replaced with the feature vectors :$x_i \xrightarrow ~\phi=\phi(x_i)$ induced by the kernel where $K_{ij}=k(x_i,x_j)=\phi(x_i)\phi(x_j)$. Hence the new predicted class label for the new example $x$ is given as :
\begin{align}
    Y^T(K+\lambda I)^{-1}k
\end{align}
Here $k=(k_1,k_2,...k_N)^T$, $k_n=x_n.x$ and $n=1,2,...N$.

\subsection{Random vector functional link network (RVFL) \cite{zhang2016comprehensive}}
RVFL \cite{zhang2016comprehensive} is the randomized version of the functional link neural network. Here, the input layer to the hidden layer, the weights are generated randomly in a suitable domain and fixed in the learning stage. Weights are generated in this manner ensuring that the activation functions $g(a_j^T x+b_j)$ are not all saturated. All weights are generated with uniform distribution within $[-S,+S]$. Here $S$ is a scale factor which is determined at the stage of parameter tuning. Only the output weights need to be determined by solving the problem :
\begin{align}
    y_i=d_i^T\beta, ~~i=1,2,...N
\end{align}
Here ${P}$ is the number of data samples, ${t}$ is the target and ${d}$ is the vectorised concatenation of random and the original features. Least squares can be used as a regularization technique in order to avoid over-fitting and obtain the solution. The two classes of RVFL algorithm are iterative RVFL, which obtains the output weights in an iterative manner based on the gradient of the error function and the closed-form based RVFL, which obtains the output weights in a single-step. The closed-form based RVFL exhibits a higher efficiency. L2 norm regularized least square is used to solve the following problem :
\begin{align}
    \sum_i (y_i -d_i^T\beta)^2 + \lambda||\beta||^2 ~;~i=1,2,\cdots,N
\end{align}
The solution for the same is given as $\beta = D(D^TD+\lambda I)^{-1} Y$, where $\lambda$ is the regularization parameter to be tuned. D and Y are the stacked features and targets of all the data samples in matrix form.

\subsection{Random vector functional link network with AutoEncoder (RVFLAE) \cite{zhang2019unsupervised}}
Autoencoder is an unsupervised learning model for which the output and input layers share the same neurons in order to reconstruct its own inputs instead of predicting target values for given input data. Sparse pre-trained RVFL is the unsupervised parameter learning method for RVFL. A sparse autoencoder is used to learn appropriate network parameters, which are developed via $l_1$ norm optimization instead of the usual $l_2$ norm. This means that more informative features will be retained to participate in the subsequent learning processes. During the learning process, the sparse autoencoder captures the excellent features in the encoding stage and learns the output weights in the decoding stage. Let the input data be ${X}$, then the sparse autoencoder has optimization problem given as : 
\begin{align}
    O_w = arg~min\{||\Tilde{H}\bar{\omega}-{X}||^2 + ||\bar{\omega}||_{l_1}\}\nonumber
\end{align}
Here $\Tilde{H} \in \mathbb{R}^{N \times L}$ is the output matrix of hidden layer obtained via random feature mapping. $\bar{\omega} \in \mathbb{R}^{L \times d} $ is the output weight matrix of the sparse encoder. $||\Tilde{H}\bar{\omega}-{X}||^2$ measures the loss to model the reconstruction process of input data and $||\bar{\omega}||_{l_1}$ is the $l_1$ norm regularization. The solution of this optimization problem is given by Fast iterative shrinkage threshold algorithm {(FISTA)\cite{beck2009fast}}.

\subsection{K nearest neighbours}
$K$ nearest neighbours algorithm assigns the label depending upon the similarity of the point with its neighbours. A constant $K$ is first chosen for the algorithm. The Euclidean distance of the given point is calculated and the K nearest members are selected from it. The number of data points is counted and assign data points is assigned to the category for which there are maximum number of neighbours.

\subsection{Neural Network}
Neural networks \cite{kingma2014adam} are network of node layers comprising of an input layer, multiple hidden layers and an output layer. Each layer has multiple number of nodes and the nodes of each layer are interconnected with the other layers. Each of the nodes consists of weights which are tuned by training on the examples. The output of each of the layer is calculated through an activation function, which is then passed to the next layer. We have divided the dataset into 85\% training and 15\% test set for our network. The layers of our network depending upon the activation functions are $\text{Feature Input Layer} \xrightarrow~ \text{Fully Connected Layer} \xrightarrow~  \text{Batch Normalized Layer}\xrightarrow ~ \text{Relu Layer}\xrightarrow~\text{Fully Connected Layer}\xrightarrow~\text{SoftMaxLayer}\xrightarrow~\text{Classification Layer}$.

For the tuning of our parameters, we have used Adam Optimizer. It is the combination of the Stochastic Gradient Descent with momentum and the Root Mean Square Propagation and hence is quite efficient.

The network once trained, is used to make predictions on the test set.


\section{Statistical Analysis}
We follow Friedman Test\cite{friedman1940comparison,friedman1937use} to test the significant difference among the linear and non-linear classification models. 
Consider $k$ algorithms are evaluated $N$ datasets/feature selection techniques. Let $r_i^j$ be the rank of $j^{th}$ algorithm on $i^{th}$ feature selection technique. The ranks of the algorithms are compared by taking their average performance $R_j = \frac{1}{N} \sum_i r_i^j$. As per the null hypotheses, the ranks $R_j$ of the algorithms should be equal considering that all algorithms are equivalent.
\begin{align*}
    \chi_F^2 = \frac{12N}{(k)(k+1)} [\sum_j R_j^2 - \frac{k(k+1)^2}{4}]
\end{align*}
Here the Friedman statistic is distributed as per $\chi_F^2$ with degrees of freedom as $k-1$ considering $k$ and $N$ are large($N>10$ and $k>5$). \\
Friedman's $\chi_F^2$ is considered to be conservative and a better statistic has been derived\cite{iman1980approximations}. The better statistic $F_F$ is given as follows
\begin{align*}
    F_F = \frac{(N-1)\chi_F^2}{N(k-1) - \chi_F^2}
\end{align*}
This statistic having $k-1$ and $(k-1)(N-1)$ degrees of freedom is distributed as per the $F$ distribution.\\
We can proceed with a post-hoc test in case the null-hypothesis is rejected. The Nemenyi test\cite{nemenyi1963distribution} is used while comparing the classifiers with each other. If the average ranks of two classifiers differ by atleast the critical difference, 
\begin{align}
    \text{Critical Difference} = q_\alpha \sqrt{\frac{k(k+1)}{6N}}
\end{align}
then the performance of classifiers is considered to be different significantly. Here, the $q_\alpha$ critical values are dependent upon Studentized range statistic divided by $\sqrt{2}$. Here, $N=7, K=12$. With simple calculation, we have $\chi^2_F=54.48, F_F=14.52$. From Statistical tables, $F_F(11,66)=1.937$. Since $14.52>1.937$, hence we reject the null hypothesis. Thus, significant difference exists among the models. With $q_{\alpha=0.05}=3.268$, we get $CD=6.3$. Hence, if the rank difference of two  classifiers is atleast $CD$, then the two models are significantly different.
Based on the Nemenyi test, Table \ref{tab:Stastical} gives the significant difference among the models. No significant difference exists among the other methods not given in Table \ref{tab:Stastical}.

\begin{table*}[!bp]
\caption{Statistical analysis on linear and non-linear variants of classification models.}
\label{tab:Stastical}
\resizebox{\textwidth}{!}{%
\begin{tabular}{|l|llllll|}
\hline
Models & KRR (Linear) & KRR (Non-Linear) & LSTWSVM (Linear) & TWSVM (Linear) & SVM (Linear) & SVM (Non-Linear) \\\hline
RELSTSVM (Non-Linear) & Yes     & No      & Yes         & No        & Yes     & Yes     \\
TBSVM (Non-Linear)    & Yes     & Yes     & Yes         & Yes       & Yes     & Yes     \\
TWSVM (Non-Linear)    & Yes     & No      & Yes         & No        & Yes     & Yes    \\\hline
\end{tabular}%
}
\vspace{-9mm}
\end{table*}

\begin{figure}[htbp]
\centering
\subfloat[White Matter]{
	\label{subfig:CF_WM}
	\includegraphics[width=0.6\textwidth]{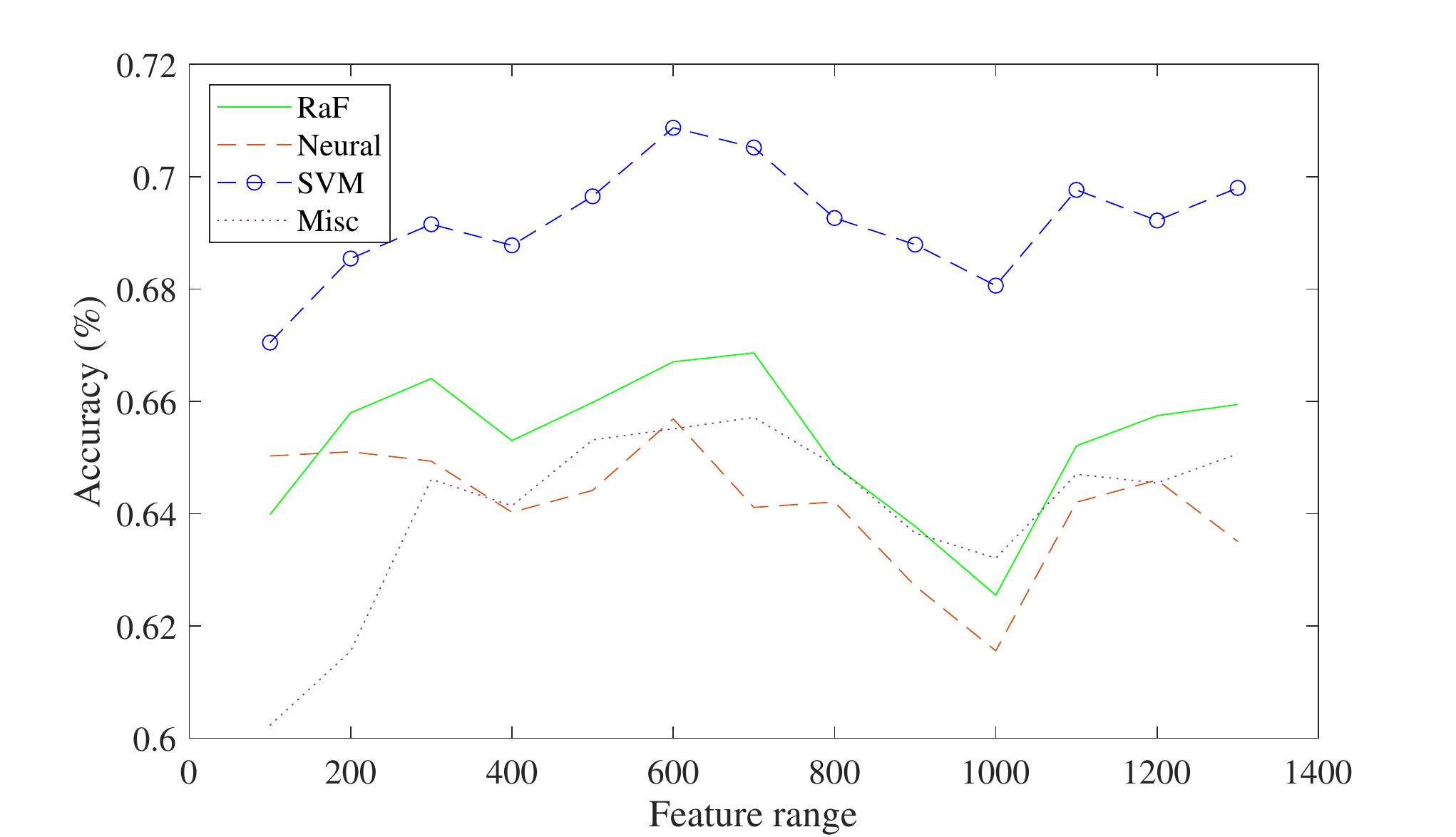}} 

\subfloat[Grey Matter]{
	\label{subfig:CF_GM}
	\includegraphics[width=0.6\textwidth]{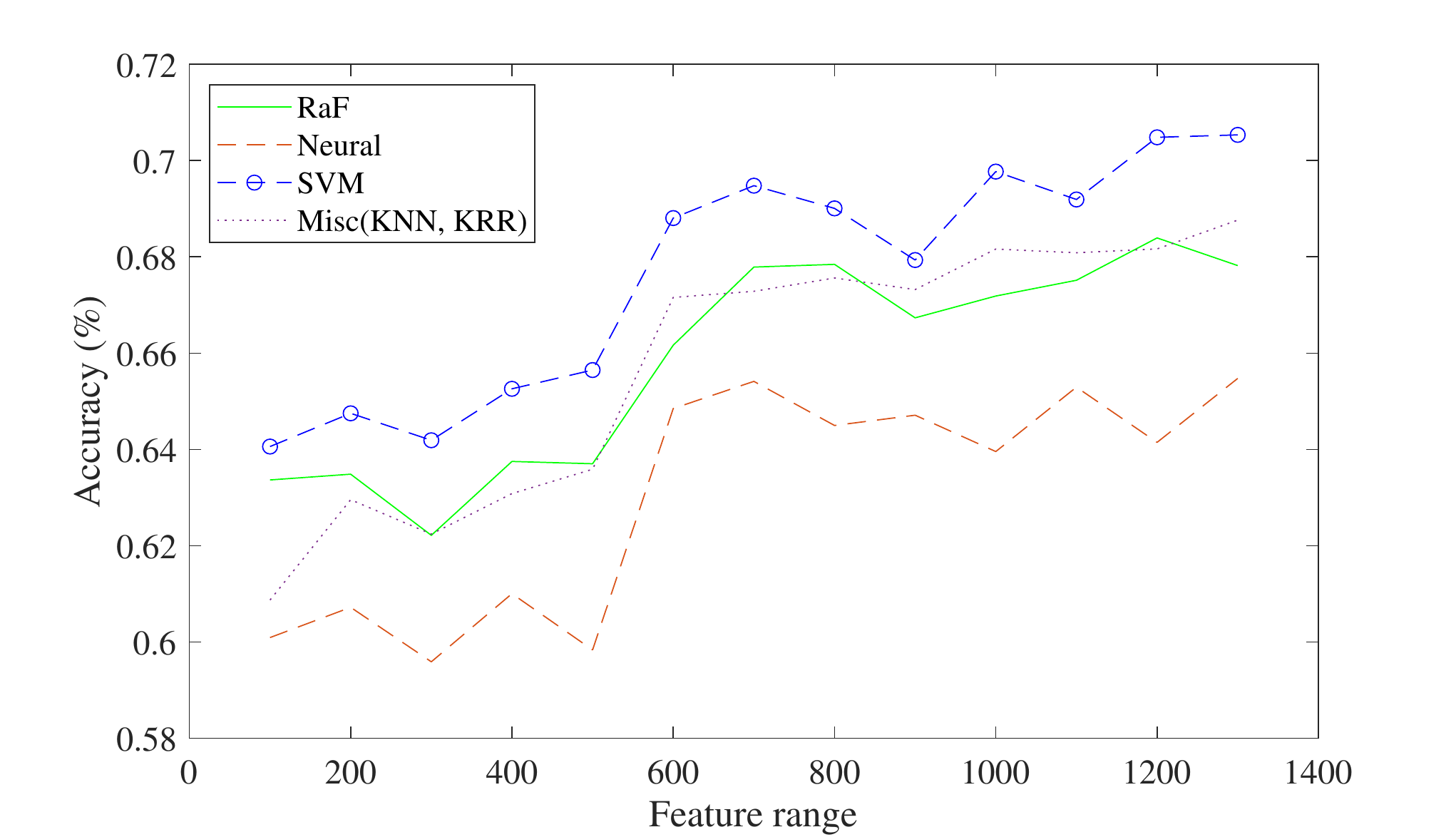} } 
	
\caption{Average performance of classifiers families}
\label{fig:classifiers_families}
\vspace{-8mm}
\end{figure}

\begin{figure}[htbp]
\centering
\subfloat[White Matter]{
	\label{subfig:FS_WM}
	\includegraphics[width=0.6\textwidth]{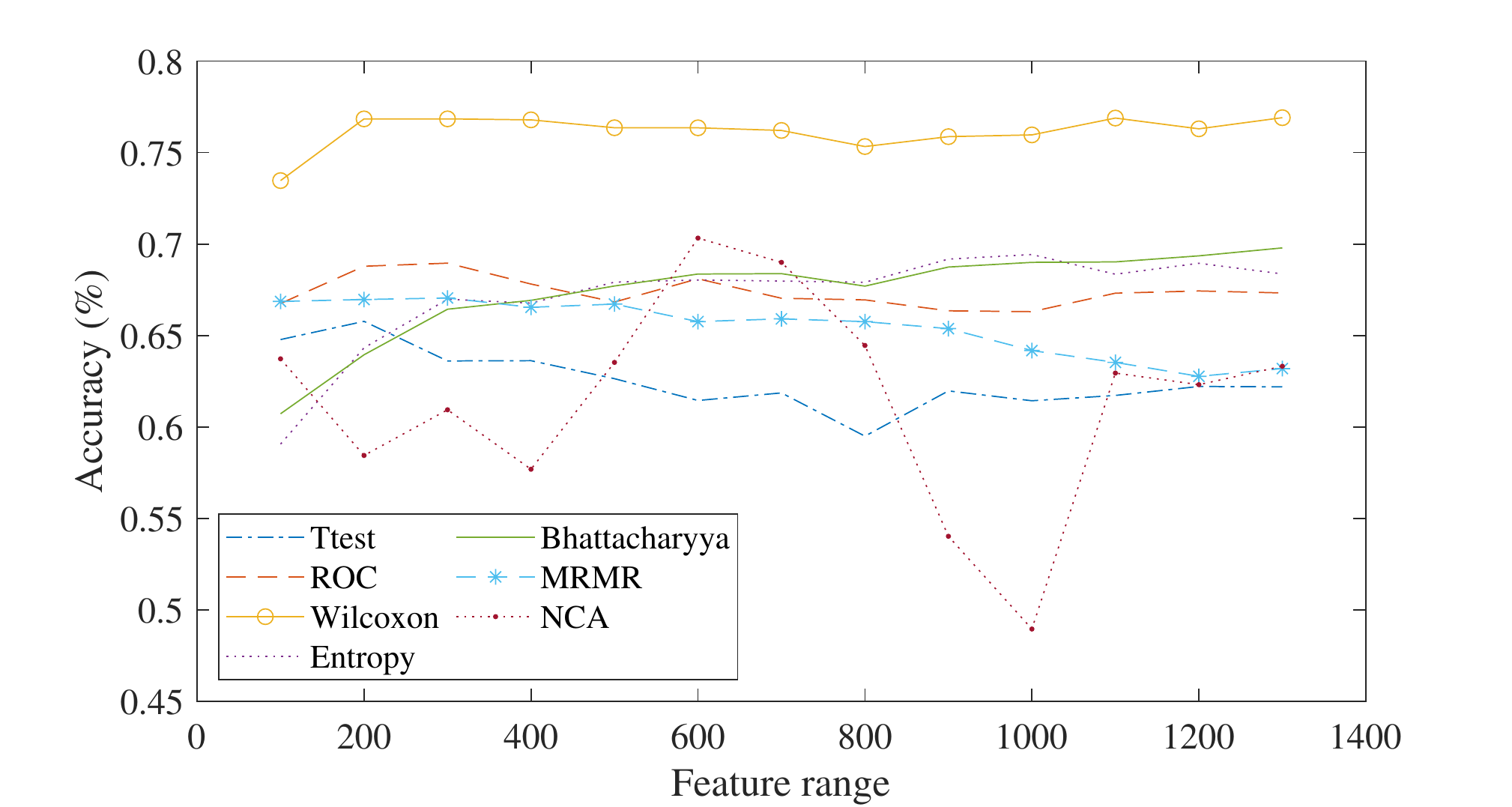}} 
	
\subfloat[Grey Matter]{
	\label{subfig:FS_GM}
	\includegraphics[width=0.6\textwidth]{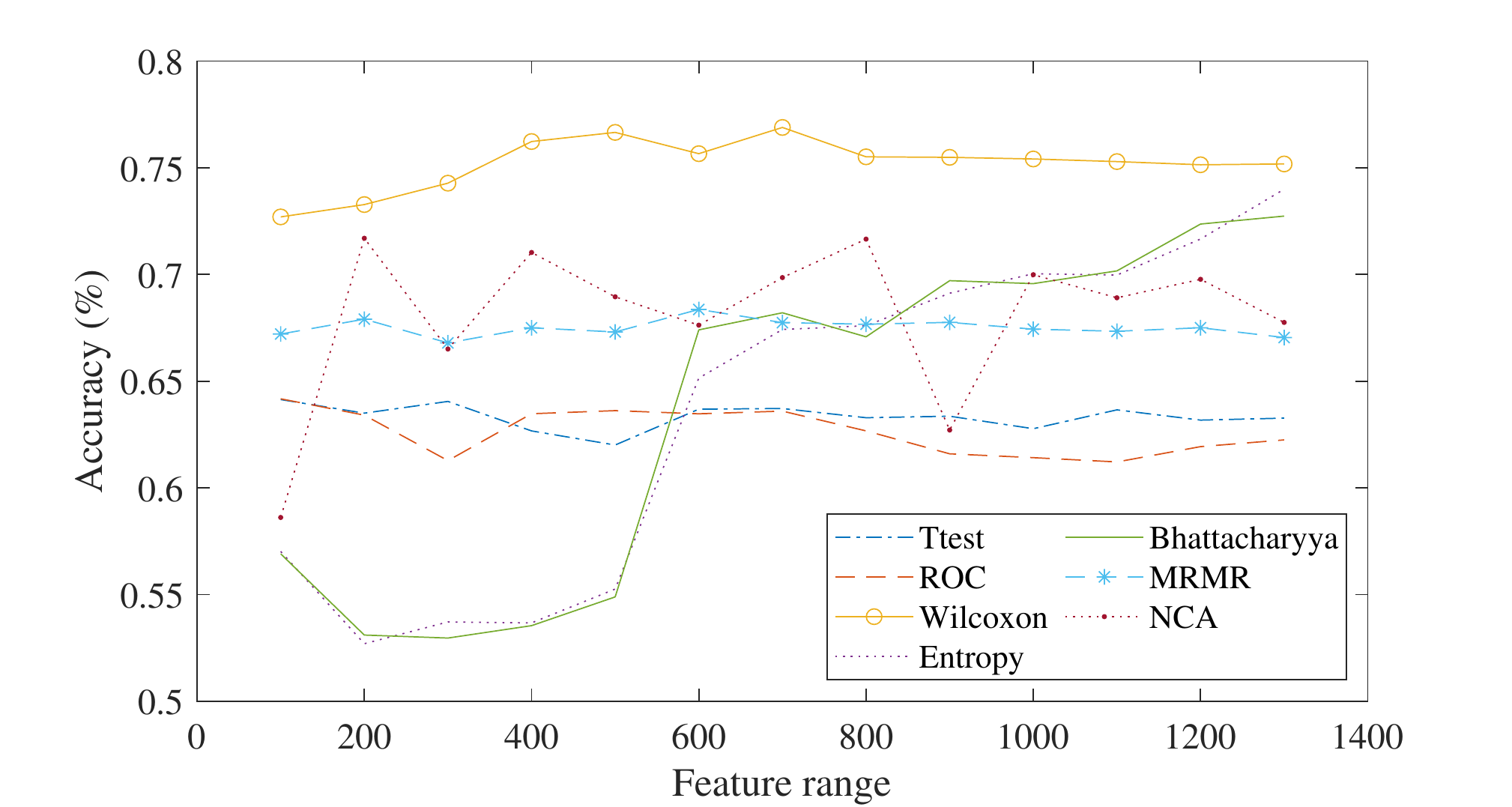} } 
	
\caption{Average performance of feature selection methods}
\label{fig:FS}
\end{figure}

\section{Supplementary Tables}
Tables \ref{table:Acc_FS_white} and \ref{table:Acc_FS_grey} are accuracies for White Matter (900 features) and Grey Matter (1200 features). Tables \ref{tab:WM1}, \ref{tab:WM2}, \ref{tab:WM3}, \ref{tab:WM4}, \ref{tab:WM5}, \ref{tab:WM6} are other performance metrics (AUCs, Sensitivity, Specificity, Precision, F-Measure and G-Mean) for White Matter (900 features). Tables \ref{tab:CM1}, \ref{tab:CM2}, \ref{tab:CM3}, \ref{tab:CM4}, \ref{tab:CM5}, \ref{tab:CM6} are other performance metrics for integrated GM and WM (500 features). Tables \ref{tab:GM1}, \ref{tab:GM2}, \ref{tab:GM3}, \ref{tab:GM4}, \ref{tab:GM5}, \ref{tab:GM6} are other performance metrics for Grey Matter (1200 features).

\begin{table}[htbp]
\caption{Accuracies for White Matter for 900 features.}
  \centering
  \tabcolsep=0.11cm
\begin{tabular}{llllllll}
 & \multicolumn{7}{c}{Feature selection methods}\\
\cmidrule(lr){2-8}
Classification &T-Test &ROC &Wilc- &Entr- &Bhatta- &MRMR &NCA\\
techniques & & &oxon &opy &charyya & &\\
\hline
\multicolumn{8}{l}{\textbf{Linear variations:}}\\
\hline
\begin{filecontents*}{WM_900_results_lin.csv}
aa,aa,aa,aa,aa,aa,aa,aa
Het-RaF,67.95,67.24,84.05,66.57,67.76,66.33,51.9
KNN,54.86,65.14,74.52,57,55.62,60.95,44.48
MPRaF-N,59.62,66.52,70.24,68.48,61.67,65,47.24
MPRaF-P,65.19,65.62,69.67,69.24,65,66.62,50.76
MPRaF-T,63.05,51.43,73.71,65.14,66.33,55.43,49.33
Neural,55.57,66.62,81.9,63.57,60.86,62.38,47
pinGTSVM,54.1,58.19,73.19,71.24,68.67,50.05,49.81
RaF-LDA,56.86,63.67,76.57,62.05,67.81,67.86,48.57
RaF-PCA,62.24,68.57,72.38,67.71,64.48,63.14,47.24
RaF,61.76,65,70.52,69.14,63,68.57,50.76
RVFLAE,50.1,58.71,72.48,60.81,58.71,60.19,56.71
RVFL,61.67,61,77.95,64.14,72.52,68.33,55.71
KRR,66.52,69.86,73.76,71,71,52,54.19
LSTWSVM,62.92,63.79,75.49,67.5,66.17,65.92,56.99
RELSTSVM,65.88,72.63,77.79,74.06,74.06,70.01,59.54
SVM,58.86,64.38,74.43,67.57,68.24,64.29,56.67
TBSVM,66.33,72,77.29,72.43,72.38,69.71,60.71
TWSVM,59.62,65.1,77.24,68.24,68.29,66.95,59.57
\end{filecontents*}
\csvreader[late after line=\\,
            late after last line=,
            before reading={\catcode`\#=12},
            after reading={\catcode`\#=6}]
            {WM_900_results_lin.csv}
            {1=\methods,2=\ttest,3=\roc,4=\wilcoxon,5=\entropy,6=\bhattacharyya,7=\MRMR,8=\NCA}
            {\methods & \ttest & \roc & \wilcoxon & \entropy & \bhattacharyya & \MRMR & \NCA}\\
\hline
\multicolumn{8}{l}{\textbf{Non-linear (Gaussian Kernel)} variations:}\\
\hline
\end{tabular}
\label{table:Acc_FS_white}
\end{table}

\begin{table}[htbp]
\caption{Accuracies for Grey Matter for 1200 features.}
  \centering
  \tabcolsep=0.11cm
\begin{tabular}{llllllll}
 & \multicolumn{7}{c}{Feature selection methods}\\
\cmidrule(lr){2-8}
Classification &T-Test &ROC &Wilc- &Entr- &Bhatta- &MRMR &NCA\\
techniques & & &oxon &opy &charyya & &\\
\hline
\multicolumn{8}{l}{\textbf{Linear variations:}}\\
\hline
\begin{filecontents*}{GM_1200_results_lin.csv}
aa,aa,aa,aa,aa,aa,aa,aa
Het-RaF,62.86,61.52,76.71,71.29,73.95,70.43,67.71
KNN,62.24,60.86,73.29,63,64.33,67.76,65.14
MPRaF-N,61.52,63.57,72.43,69.62,74.43,67.62,71.24
MPRaF-P,65,63.57,69.71,71.86,71,68.95,69.14
MPRaF-T,67.76,64.9,75.9,69.52,70.24,64.29,71.81
Neural,62.24,60.81,74.62,61.57,65,61.67,66.38
pinGTSVM,56.76,53.29,71.76,67.05,69.86,62.14,66.38
RaF-LDA,65.81,56.71,77.9,69.9,65,67.71,66.19
RaF-PCA,67,58.62,77.24,68.95,73.76,67.81,61.71
RaF,66.29,63.62,71,70.29,71.14,68.29,67.81
RVFLAE,54.86,53.43,69.71,58.38,61.76,60.38,64.29
RVFL,61.67,65.48,71.81,72.52,67,68.52,65
KRR,60.86,58.76,76.67,70.52,70.48,67.67,70.38
LSTWSVM,59.5,58.25,75.26,70.3,69.89,67.94,69.56
RELSTSVM,64.9,64.26,78.26,73.73,73.33,69.75,73.91
SVM,57.62,52.76,74.57,71.86,73.24,67.1,71.14
TBSVM,64.24,64.29,79.48,72.48,73.9,69.29,71.81
TWSVM,60.24,58.95,74.57,72.48,67.86,67.05,66.48
\end{filecontents*}
\csvreader[late after line=\\,
            late after last line=,
            before reading={\catcode`\#=12},
            after reading={\catcode`\#=6}]
            {GM_1200_results_lin.csv}
            {1=\methods,2=\ttest,3=\roc,4=\wilcoxon,5=\entropy,6=\bhattacharyya,7=\MRMR,8=\NCA}
            {\methods & \ttest & \roc & \wilcoxon & \entropy & \bhattacharyya & \MRMR & \NCA}\\
\hline
\multicolumn{8}{l}{\textbf{Non-linear (Gaussian Kernel) variations:}}\\
\hline
\end{tabular}
\label{table:Acc_FS_grey}
\end{table}

\begin{table}[htbp]
\caption{Evaluation of classification models based on AUCs with White Matter (900 features).}
\label{tab:WM1}
\resizebox{\textwidth}{!}{%
\begin{tabular}{|l|lllllll|}
\hline
Methods              & T-Test & ROC   & Wilcoxon & Entropy & Bhattacharyya & MRMR  & NCA   \\\hline
Het-RaF               & 67.98  & 67.29 & 83.68    & 68.51   & 69.14         & 67.58 & 51.23 \\
KNN                   & 54.06  & 65.40 & 74.88    & 56.24   & 54.97         & 61.32 & 45.83 \\
KRR (Linear)          & 65.89  & 70.42 & 73.66    & 72.47   & 72.47         & 55.03 & 54.55 \\
KRR (Non-Linear)      & 65.33  & 67.42 & 73.99    & 71.09   & 71.16         & 68.72 & 53.91 \\
LSTWSVM (Linear)      & 57.12  & 59.69 & 68.22    & 64.65   & 63.65         & 57.70 & 55.08 \\
LSTWSVM (Non-Linear)  & 60.46  & 68.59 & 88.54    & 81.65   & 78.00         & 65.92 & 51.61 \\
MPRaF-N               & 63.58  & 67.71 & 73.03    & 70.68   & 64.32         & 66.26 & 51.37 \\
\hline
\end{tabular}%
}
\end{table}

\begin{table}[htbp]
\caption{Sensitivities of the classification models for White Matter (900 features).}
\label{tab:WM2}
\resizebox{\textwidth}{!}{%
\begin{tabular}{|l|lllllll|}
\hline
Methods              & T-Test & ROC   & Wilcoxon & Entropy & Bhattacharyya & MRMR  & NCA   \\\hline
Het-RaF               & 63.27  & 63.56 & 80.56    & 59.99   & 59.88         & 60.84 & 38.35 \\
KNN                   & 25.35  & 52.64 & 62.98    & 17.20   & 16.09         & 41.32 & 18.92 \\
KRR (Linear)          & 65.13  & 69.07 & 72.19    & 70.23   & 70.23         & 67.75 & 52.38 \\
KRR (Non-Linear)      & 53.96  & 70.05 & 61.21    & 61.12   & 60.00         & 62.29 & 30.03 \\
LSTWSVM (Linear)      & 68.73  & 67.88 & 82.75    & 70.35   & 68.69         & 74.15 & 58.91 \\
LSTWSVM (Non-Linear)  & 73.35  & 73.56 & 70.04    & 73.62   & 80.81         & 77.19 & 65.20 \\
MPRaF-N               & 76.25  & 68.80 & 82.48    & 63.42   & 57.25         & 65.19 & 67.48 \\
\hline
\end{tabular}%
}
\end{table}

\begin{table}[htbp]
\caption{Specificity of classification models for White Matter (900 features).}
\label{tab:WM3}
\resizebox{\textwidth}{!}{%
\begin{tabular}{|l|lllllll|}
\hline
Methods              & T-Test & ROC   & Wilcoxon & Entropy & Bhattacharyya & MRMR  & NCA   \\\hline
Het-RaF               & 72.70  & 71.02 & 86.80    & 77.04   & 78.40         & 74.33 & 64.11 \\
KNN                   & 82.77  & 78.16 & 86.77    & 95.28   & 93.85         & 81.33 & 72.73 \\
KRR (Linear)          & 66.65  & 71.77 & 75.14    & 74.70   & 74.70         & 42.32 & 56.73 \\
KRR (Non-Linear)      & 76.69  & 64.79 & 86.78    & 81.06   & 82.31         & 75.14 & 77.80 \\
LSTWSVM (Linear)      & 67.19  & 65.39 & 77.44    & 66.78   & 65.42         & 71.14 & 57.96 \\
LSTWSVM (Non-Linear)  & 72.38  & 72.41 & 73.68    & 74.09   & 79.13         & 71.69 & 57.39 \\
MPRaF-N               & 50.90  & 66.62 & 63.59    & 77.95   & 71.38         & 67.33 & 35.25 \\
\hline
\end{tabular}%
}
\end{table}

\begin{table}[htbp]
\caption{Precision of classification models for White Matter (900 features).}
\label{tab:WM4}
\resizebox{\textwidth}{!}{%
\begin{tabular}{|l|lllllll|}
\hline
Methods              & T-Test & ROC   & Wilcoxon & Entropy & Bhattacharyya & MRMR  & NCA   \\\hline
Het-RaF               & 69.45  & 69.61 & 84.71    & 74.13   & 73.58         & 69.92 & 50.08 \\
KNN                   & 52.83  & 70.18 & 81.75    & NaN     & NaN           & 67.98 & 32.95 \\
KRR (Linear)          & 69.73  & 70.97 & 72.06    & 71.31   & 71.31         & 55.01 & 53.35 \\
KRR (Non-Linear)      & 71.87  & 67.75 & 79.48    & 77.53   & 78.89         & 69.09 & 64.19 \\
LSTWSVM (Linear)      & 59.52  & 60.49 & 71.45    & 64.78   & 63.67         & 61.82 & 54.45 \\
LSTWSVM (Non-Linear)  & 63.24  & 68.73 & 79.42    & 76.75   & 75.78         & 67.13 & 52.89 \\
MPRaF-N               & 63.51  & 68.59 & 67.60    & 75.48   & 69.35         & 66.65 & 50.71 \\
\hline
\end{tabular}%
}
\end{table}

\begin{table}[htbp]
\caption{F-Measures for White Matter for 900 features.}
\label{tab:WM5}
\resizebox{\textwidth}{!}{%
\begin{tabular}{|l|lllllll|}
\hline
Methods               & T-Test & ROC   & Wilcoxon & Entropy & Bhattacharyya & MRMR  & NCA   \\\hline
Het-RaF               & 64.33  & 64.11 & 81.89    & 61.47   & 62.45         & 62.39 & NaN   \\
KNN                   & NaN    & 58.40 & 69.46    & NaN     & NaN           & 47.93 & NaN   \\
KRR (Linear)          & 65.47  & 68.22 & 71.02    & 68.54   & 68.54         & 57.01 & 51.34 \\
KRR (Non-Linear)      & 58.46  & 66.42 & 68.33    & 65.49   & 65.29         & 64.73 & 37.97 \\
LSTWSVM (Linear)      & 60.79  & 61.48 & 72.13    & 65.25   & 64.10         & 63.07 & 55.46 \\
LSTWSVM (Non-Linear)  & 64.74  & 69.60 & 80.26    & 77.30   & 77.14         & 67.93 & 53.68 \\
MPRaF-N               & 63.96  & 66.09 & 71.52    & 65.35   & 56.24         & 63.49 & 54.44 \\
\hline
\end{tabular}%
}
\end{table}

\begin{table}[htbp]
\caption{G-Means for White Matter for 900 features.}
\label{tab:WM6}
\resizebox{\textwidth}{!}{%
\begin{tabular}{|l|lllllll|}
\hline
Methods              & T-Test & ROC   & Wilcoxon & Entropy & Bhattacharyya & MRMR  & NCA   \\\hline
Het-RaF               & 65.31  & 65.26 & 82.26    & 63.72   & 64.35         & 63.79 & 43.29 \\
KNN                   & 35.86  & 59.82 & 70.88    & NaN     & NaN           & 50.71 & 22.64 \\
KRR (Linear)          & 66.43  & 69.10 & 71.57    & 69.62   & 69.62         & 59.09 & 52.08 \\
KRR (Non-Linear)      & 60.47  & 67.62 & 69.32    & 67.29   & 67.24         & 65.21 & 41.85 \\
LSTWSVM (Linear)      & 5.61   & 5.57  & 5.56     & 5.52    & 5.75          & 5.72  & 5.51  \\
LSTWSVM (Non-Linear)  & 0.31   & 0.29  & 0.31     & 0.31    & 0.31          & 0.31  & 0.31  \\
MPRaF-N               & 66.75  & 67.35 & 73.24    & 67.27   & 59.14         & 64.66 & 56.64 \\
\hline
\end{tabular}%
}
\end{table}

\begin{table}[htbp]
\caption{AUCs for Combined Matter for 500 features.}
\label{tab:CM1}
\resizebox{\textwidth}{!}{%
\begin{tabular}{|l|lllllll|}
\hline
Methods               & T-Test & ROC   & Wilcoxon & Entropy & Bhattacharyya & MRMR  & NCA   \\\hline
Het-RaF               & 72.16  & 73.21 & 81.50    & 69.35   & 63.93         & 73.62 & 64.08 \\
KNN                   & 68.77  & 71.92 & 72.53    & 59.92   & 59.92         & 65.27 & 69.26 \\
KRR (Linear)          & 72.22  & 71.00 & 74.07    & 63.64   & 64.98         & 73.74 & 66.33 \\
KRR (Non-Linear)      & 71.89  & 73.68 & 72.18    & 66.68   & 67.73         & 73.22 & 67.17 \\
LSTWSVM (Linear)      & 68.44  & 64.33 & 68.77    & 65.21   & 63.21         & 66.38 & 66.04 \\
LSTWSVM (Non-Linear)  & 67.99  & 64.88 & 90.13    & 72.05   & 70.21         & 69.51 & 64.70 \\
MPRaF-N               & 73.17  & 72.11 & 73.35    & 65.75   & 63.99         & 67.87 & 63.09 \\
\hline
\end{tabular}%
}
\end{table}

\begin{table}[htbp]
\caption{Sensitivities for Combined Matter for 500 features.}
\label{tab:CM2}
\resizebox{\textwidth}{!}{%
\begin{tabular}{|l|lllllll|}
\hline
Methods               & T-Test & ROC   & Wilcoxon & Entropy & Bhattacharyya & MRMR  & NCA   \\\hline
Het-RaF               & 61.14  & 64.89 & 80.55    & 58.04   & 53.74         & 70.66 & 52.66 \\
KNN                   & 58.22  & 67.94 & 67.56    & 26.00   & 26.00         & 53.73 & 56.13 \\
KRR (Linear)          & 71.35  & 65.93 & 70.08    & 59.34   & 60.77         & 71.70 & 62.32 \\
KRR (Non-Linear)      & 66.47  & 65.38 & 55.99    & 50.60   & 54.95         & 65.70 & 56.52 \\
LSTWSVM (Linear)      & 75.26  & 66.90 & 82.04    & 73.35   & 73.29         & 68.29 & 61.38 \\
LSTWSVM (Non-Linear)  & 81.21  & 84.16 & 69.55    & 84.71   & 83.60         & 77.01 & 74.79 \\
MPRaF-N               & 77.04  & 75.87 & 74.59    & 47.13   & 48.89         & 63.74 & 60.28 \\
\hline
\end{tabular}%
}
\end{table}

\begin{table}[htbp]
\caption{Specificity's for Combined Matter for 500 features.}
\label{tab:CM3}
\resizebox{\textwidth}{!}{%
\begin{tabular}{|l|lllllll|}
\hline
Methods               & T-Test & ROC   & Wilcoxon & Entropy & Bhattacharyya & MRMR  & NCA   \\\hline
Het-RaF               & 83.19  & 81.53 & 82.45    & 80.65   & 74.12         & 76.58 & 75.50 \\
KNN                   & 79.33  & 75.91 & 77.49    & 93.85   & 93.85         & 76.80 & 82.38 \\
KRR (Linear)          & 73.09  & 76.06 & 78.06    & 67.95   & 69.20         & 75.77 & 70.35 \\
KRR (Non-Linear)      & 77.31  & 81.99 & 88.37    & 82.76   & 80.51         & 80.74 & 77.82 \\
LSTWSVM (Linear)      & 74.46  & 64.95 & 79.68    & 71.10   & 71.17         & 67.50 & 63.50 \\
LSTWSVM (Non-Linear)  & 79.17  & 80.32 & 73.93    & 80.24   & 78.04         & 73.64 & 69.01 \\
MPRaF-N               & 69.30  & 68.34 & 72.10    & 84.38   & 79.08         & 72.00 & 65.90 \\
\hline
\end{tabular}%
}
\end{table}

\begin{table}[htbp]
\caption{Precisions for Combined Matter for 500 features.}
\label{tab:CM4}
\resizebox{\textwidth}{!}{%
\begin{tabular}{|l|lllllll|}
\hline
Methods               & T-Test & ROC   & Wilcoxon & Entropy & Bhattacharyya & MRMR  & NCA   \\\hline
Het-RaF               & 75.62  & 75.06 & 81.49    & 75.64   & 70.40         & 75.96 & 65.13 \\
KNN                   & 74.84  & 71.14 & 77.50    & 75.83   & 75.83         & 69.92 & 75.78 \\
KRR (Linear)          & 69.43  & 72.86 & 76.24    & 64.59   & 66.25         & 72.92 & 63.96 \\
KRR (Non-Linear)      & 73.21  & 79.03 & 86.25    & 77.51   & 75.37         & 78.31 & 71.71 \\
LSTWSVM (Linear)      & 69.00  & 63.07 & 72.29    & 64.18   & 62.74         & 65.72 & 63.80 \\
LSTWSVM (Non-Linear)  & 70.68  & 69.05 & 80.06    & 74.73   & 72.87         & 69.52 & 64.72 \\
MPRaF-N               & 67.58  & 68.98 & 70.85    & 75.31   & 74.68         & 69.81 & 62.92 \\
\hline
\end{tabular}%
}
\end{table}

\begin{table}[htbp]
\caption{F-Measures for Combined Matter for 500 features.}
\label{tab:CM5}
\resizebox{\textwidth}{!}{%
\begin{tabular}{|l|lllllll|}
\hline
Methods               & T-Test & ROC   & Wilcoxon & Entropy & Bhattacharyya & MRMR  & NCA   \\\hline
Het-RaF               & 66.34  & 68.04 & 80.07    & 62.25   & 56.58         & 71.36 & 55.72 \\
KNN                   & 64.40  & 67.92 & 69.98    & NaN     & NaN           & 58.93 & 61.68 \\
KRR (Linear)          & 69.55  & 67.96 & 71.66    & 59.90   & 61.44         & 70.89 & 61.19 \\
KRR (Non-Linear)      & 68.76  & 69.24 & 65.33    & 58.08   & 60.13         & 69.29 & 61.39 \\
LSTWSVM (Linear)      & 70.20  & 63.84 & 73.25    & 66.09   & 64.82         & 66.32 & 64.28 \\
LSTWSVM (Non-Linear)  & 72.08  & 70.76 & 81.03    & 75.43   & 73.49         & 70.54 & 65.75 \\
MPRaF-N               & 70.78  & 70.60 & 71.45    & 53.91   & 55.92         & 65.11 & 59.20 \\
\hline
\end{tabular}%
}
\end{table}

\begin{table}[htbp]
\caption{G-Means for Combined Matter for 500 features.}
\label{tab:CM6}
\resizebox{\textwidth}{!}{%
\begin{tabular}{|l|lllllll|}
\hline
Methods               & T-Test & ROC   & Wilcoxon & Entropy & Bhattacharyya & MRMR  & NCA   \\\hline
Het-RaF               & 67.34  & 68.97 & 80.54    & 64.04   & 58.81         & 72.32 & 57.22 \\
KNN                   & 65.44  & 68.72 & 71.19    & 43.47   & 43.47         & 60.29 & 63.71 \\
KRR (Linear)          & 69.97  & 68.66 & 72.39    & 60.90   & 62.44         & 71.59 & 62.10 \\
KRR (Non-Linear)      & 69.29  & 70.67 & 68.05    & 60.54   & 62.13         & 70.60 & 62.71 \\
LSTWSVM (Linear)      & 7.19   & 7.06  & 7.00     & 7.08    & 7.05          & 7.15  & 7.77  \\
LSTWSVM (Non-Linear)  & 0.47   & 0.47  & 0.49     & 0.48    & 0.48          & 0.48  & 0.48  \\
MPRaF-N               & 71.53  & 71.50 & 72.08    & 56.93   & 58.63         & 65.93 & 60.36 \\
\hline
\end{tabular}%
}
\end{table}

\begin{table}[htbp]
\caption{AUCs for Grey Matter for 1200 features.}
\label{tab:GM1}
\resizebox{\textwidth}{!}{%
\begin{tabular}{|l|lllllll|}
\hline
Methods               & T-Test & ROC   & Wilcoxon & Entropy & Bhattacharyya & MRMR  & NCA   \\\hline
Het-RaF               & 63.57  & 60.49 & 76.64    & 71.23   & 74.30         & 70.01 & 69.46 \\
KNN                   & 59.69  & 58.61 & 73.18    & 63.71   & 65.26         & 68.16 & 66.20 \\
KRR (Linear)          & 60.88  & 58.58 & 76.71    & 71.00   & 70.38         & 67.07 & 71.53 \\
KRR (Non-Linear)      & 67.94  & 68.73 & 76.21    & 74.02   & 73.18         & 70.22 & 68.02 \\
LSTWSVM (Linear)      & 58.44  & 62.63 & 75.85    & 66.52   & 63.27         & 65.87 & 65.50 \\
LSTWSVM (Non-Linear)  & 55.74  & 54.45 & 73.97    & 81.13   & 81.55         & 66.29 & 70.75 \\
MPRaF-N               & 62.84  & 63.68 & 73.53    & 70.02   & 75.02         & 66.76 & 73.31 \\
\hline
\end{tabular}%
}
\end{table}

\begin{table}[htbp]
\caption{Sensitivities for Grey Matter for 1200 features.}
\label{tab:GM2}
\resizebox{\textwidth}{!}{%
\begin{tabular}{|l|lllllll|}
\hline
Methods               & T-Test & ROC   & Wilcoxon & Entropy & Bhattacharyya & MRMR  & NCA   \\\hline
Het-RaF               & 54.57  & 55.64 & 72.07    & 66.58   & 69.92         & 63.90 & 60.97 \\
KNN                   & 46.15  & 46.00 & 66.23    & 34.75   & 37.84         & 59.15 & 37.56 \\
KRR (Linear)          & 54.17  & 53.58 & 72.31    & 59.50   & 59.92         & 61.63 & 67.98 \\
KRR (Non-Linear)      & 45.77  & 51.84 & 60.72    & 51.81   & 52.81         & 64.54 & 54.78 \\
LSTWSVM (Linear)      & 60.56  & 53.87 & 74.67    & 74.08   & 76.50         & 70.01 & 73.63 \\
LSTWSVM (Non-Linear)  & 73.31  & 75.38 & 82.35    & 80.67   & 83.46         & 75.95 & 81.84 \\
MPRaF-N               & 62.53  & 54.53 & 73.58    & 58.56   & 65.58         & 59.18 & 67.86 \\
\hline
\end{tabular}%
}
\end{table}

\begin{table}[htbp]
\caption{Specificity's for Grey Matter for 1200 features.}
\label{tab:GM3}
\resizebox{\textwidth}{!}{%
\begin{tabular}{|l|lllllll|}
\hline
Methods               & T-Test & ROC   & Wilcoxon & Entropy & Bhattacharyya & MRMR  & NCA   \\\hline
Het-RaF               & 72.56  & 65.33 & 81.21    & 75.88   & 78.69         & 76.12 & 77.95 \\
KNN                   & 73.24  & 71.21 & 80.13    & 92.68   & 92.68         & 77.17 & 94.83 \\
KRR (Linear)          & 67.58  & 63.59 & 81.12    & 82.50   & 80.83         & 72.52 & 75.08 \\
KRR (Non-Linear)      & 90.11  & 85.61 & 91.70    & 96.22   & 93.56         & 75.89 & 81.25 \\
LSTWSVM (Linear)      & 58.87  & 56.62 & 75.17    & 72.00   & 73.86         & 66.39 & 72.68 \\
LSTWSVM (Non-Linear)  & 65.85  & 70.20 & 80.73    & 80.21   & 83.99         & 77.89 & 79.83 \\
MPRaF-N               & 63.15  & 72.82 & 73.48    & 81.48   & 84.46         & 74.34 & 78.75 \\
\hline
\end{tabular}%
}
\end{table}

\begin{table}[htbp]
\caption{Precisions for Grey Matter for 1200 features.}
\label{tab:GM4}
\resizebox{\textwidth}{!}{%
\begin{tabular}{|l|lllllll|}
\hline
Methods               & T-Test & ROC   & Wilcoxon & Entropy & Bhattacharyya & MRMR  & NCA   \\\hline
Het-RaF               & 68.39  & 62.74 & 77.72    & 72.17   & 76.19         & 72.81 & 74.44 \\
KNN                   & 70.13  & 66.25 & 77.86    & 83.83   & 83.50         & 69.95 & 88.00 \\
KRR (Linear)          & 60.49  & 57.71 & 78.17    & 75.13   & 74.43         & 68.59 & 73.82 \\
KRR (Non-Linear)      & 84.74  & 81.06 & 89.31    & 92.50   & 88.33         & 73.06 & 74.64 \\
LSTWSVM (Linear)      & 56.76  & 57.24 & 74.15    & 67.33   & 65.93         & NaN   & 66.24 \\
LSTWSVM (Non-Linear)  & 59.57  & 59.09 & 76.17    & 79.34   & 80.65         & 67.14 & 73.47 \\
MPRaF-N               & 63.22  & 68.62 & 74.17    & 76.95   & 84.38         & 74.10 & 74.70 \\
\hline
\end{tabular}%
}
\end{table}

\begin{table}[htbp]
\caption{F-Measures for Grey Matter for 1200 features.}
\label{tab:GM5}
\resizebox{\textwidth}{!}{%
\begin{tabular}{|l|lllllll|}
\hline
Methods               & T-Test & ROC   & Wilcoxon & Entropy & Bhattacharyya & MRMR  & NCA   \\\hline
Het-RaF               & 57.93  & 57.06 & 73.10    & 68.43   & 71.01         & 66.81 & 63.71 \\
KNN                   & 52.62  & 52.10 & 70.19    & 46.47   & 48.91         & 61.96 & 50.05 \\
KRR (Linear)          & 55.32  & 53.49 & 73.79    & 64.42   & 64.66         & 62.90 & 67.68 \\
KRR (Non-Linear)      & 58.07  & 61.82 & 69.89    & 65.41   & 65.34         & 66.59 & 61.71 \\
LSTWSVM (Linear)      & 57.67  & 58.37 & 74.82    & 68.28   & 67.23         & 65.31 & 67.61 \\
LSTWSVM (Non-Linear)  & 60.18  & 60.61 & 76.75    & 80.00   & 81.69         & 69.51 & 74.37 \\
MPRaF-N               & 60.65  & 58.57 & 71.17    & 64.23   & 70.69         & 62.07 & 68.81 \\
\hline
\end{tabular}%
}
\end{table}

\begin{table}[htbp]
\caption{G-Means for Grey Matter for 1200 features.}
\label{tab:GM6}
\resizebox{\textwidth}{!}{%
\begin{tabular}{|l|lllllll|}
\hline
Methods               & T-Test & ROC   & Wilcoxon & Entropy & Bhattacharyya & MRMR  & NCA   \\\hline
Het-RaF               & 59.63  & 58.10 & 73.98    & 68.90   & 72.02         & 67.57 & 65.64 \\
KNN                   & 55.02  & 53.96 & 71.11    & 52.11   & 54.07         & 63.22 & 55.86 \\
KRR (Linear)          & 56.29  & 54.53 & 74.50    & 65.84   & 65.89         & 63.98 & 69.24 \\
KRR (Non-Linear)      & 61.50  & 64.04 & 72.36    & 68.63   & 67.88         & 67.68 & 63.17 \\
LSTWSVM (Linear)      & 10.42  & 10.49 & 10.41    & 10.56   & 10.49         & 10.56 & 10.55 \\
LSTWSVM (Non-Linear)  & 0.54   & 0.55  & 0.53     & 0.55    & 0.56          & 0.55  & 0.54  \\
MPRaF-N               & 61.72  & 60.03 & 72.50    & 65.95   & 72.75         & 64.08 & 70.02 \\
\hline
\end{tabular}%
}
\end{table}

\bibliographystyle{elsarticle-num}
\bibliography{refs}